\documentclass[accepted]{article}

\usepackage{microtype}
\usepackage{graphicx}
\usepackage{subfigure}
\usepackage{booktabs} 
\usepackage[whole]{bxcjkjatype} 
\usepackage{amsmath}
\usepackage{multicol}

\usepackage{hyperref}

\usepackage{mlsys2024}
\usepackage[table]{xcolor}
\usepackage{colortbl}
\definecolor{lightred}{RGB}{255, 204, 204}
\definecolor{lightyellow}{RGB}{255, 255, 204}
\definecolor{lightgreen}{RGB}{204, 255, 204}

\mlsystitlerunning{Accelerating Large Language Model with Memory Consumption Estimator}

\begin{document}

\twocolumn[
\mlsystitle{Accelerating Large Language Model Training with 4D Parallelism and Memory Consumption Estimator}



\mlsyssetsymbol{equal}{*}

\begin{mlsysauthorlist}
\mlsysauthor{Kazuki Fujii}{ed}
\mlsysauthor{Kohei Watanabe}{tu}
\mlsysauthor{Rio Yokota}{sc}
\end{mlsysauthorlist}

\mlsysaffiliation{tu}{Turing. inc, Tokyo, Japan}
\mlsysaffiliation{ed}{School of Computing, Institute of Science Tokyo, Tokyo, Japan}
\mlsysaffiliation{sc}{Super Computing Research Center, Institute of Science Tokyo, Tokyo, Japan}

\mlsyscorrespondingauthor{Kazuki Fujii}{kazuki.fujii@rio.scrc.iir.isct.ac.jp}

\mlsyskeywords{Machine Learning, MLSys}

\vskip 0.3in

\begin{abstract}
In large language model (LLM) training, several parallelization strategies, including Tensor Parallelism (TP), Pipeline Parallelism (PP), Data Parallelism (DP), as well as Sequence Parallelism (SP) and Context Parallelism (CP), are employed to distribute model parameters, activations, and optimizer states across devices. Identifying the optimal parallelization configuration for each environment while avoiding GPU memory overflow remains a challenging task. In this study, we provide precise formulas to estimate the memory consumed by parameters, gradients, optimizer states, and activations for 4D parallel training (DP, TP, PP, CP) in the Llama architecture. We conducted 454 experiments on A100 and H100 GPUs, incorporating often neglected factors such as temporary buffers and memory fragmentation into our analysis. Results indicate that when the estimated memory usage is below 80\% of the available GPU memory, the training never encounters out-of-memory errors. This simple yet effective formula allows us to identify parallelization configurations that could lead to memory overflow in advance, significantly reducing the configuration search space. Additionally, through a comprehensive exploration of optimal configurations in 4D parallelism, our analysis of the 454 experimental results provides empirical insights into optimal 4D parallelism configurations.
\end{abstract}
]



\printAffiliationsAndNotice{} 

\section{Introduction}

Training large language models (LLMs) requires model parallelism to distribute the model's parameters, optimizer states, and activations across devices to fit within GPU memory constraints.
Data parallelism is also employed to enable training within a realistic time.
Efficient LLM training utilizes not only Tensor Parallelism (TP)~\cite{mesh-tensorflow, megatron, GSPMD}, Pipeline Parallelism (PP)~\cite{gpipe,terapipe,pipedream}, and Data Parallelism (DP), but also Sequence Parallelism (SP)~\cite{korthikanti2023reducing} and Context Parallelism (CP)~\cite{context_parallel, liu2023ring}. 
However, determining the optimal parallelism configuration for a given environment while avoiding GPU memory overflow is challenging.

Excessive amounts of tensor parallelism can lead to smaller matrix multiplications, which reduce GPU utilization and degrade training throughput, as measured in TFLOP/s. 
Similarly, an oversized pipeline parallel configuration can increase pipeline bubbles, further reducing TFLOP/s. 
Additionally, when the global batch size is fixed, scaling the number of GPUs and employing pipeline parallelism while increasing the data parallel size reduces the number of microbatches, making pipeline bubbles prominent and causing performance degradation beyond the communication overhead of added GPUs. Consequently, identifying an optimal parallelism configuration is non-trivial.

Moreover, when utilizing supercomputers with GPU memory capacities and per-node GPU counts different from those typically offered by cloud services—such as the NVIDIA H100 (SXM) 94GB×4 that we used in our experiments-one cannot reuse configurations that were optimal in common environments like H100 80GB×8.
For example, even if setting TP=8 is optimal on an H100×8 setup, when using a cluster configured with H100 94GB×4, specifying a tensor parallel size exceeding the per-node GPU count of 4 leads to significant training speed degradation due to the additional communication cost introduced by tensor parallelism. Additionally, configurations that result in out-of-memory with 80GB memory may be trainable with 94GB memory.
Therefore, a setting optimized in one environment does not necessarily apply to another. However, indiscriminately sweeping all parallelism configurations for each training environment requires substantial computational cost and time.

In the training of Llama 3~\cite{dubey2024llama3herdmodels}, they addressed this issue by developing a memory consumption estimator, but its implementation has not been made public, and the open community cannot benefit from it.
ZeRO~\cite{zero} provides equations to calculate the memory size consumed by parameters, gradients, and optimizer states, but these equations are limited to the GPT architecture, and the calculation for activation memory remains approximate, posing practical difficulties.
Other prior work~\cite{korthikanti2023reducing} presented equations for computing activations in GPT models, but these equations are for the case of 3D parallelism (DP, TP, PP) with sequence parallelism (SP), and have practical issues such as lack of support for the Llama architecture, which has been widely adopted recently, nor do they consider context parallelism, which is essential for handling long contexts.

In this work, we provide equations for the memory consumption of parameters, gradients, optimizer states, and activations when training with 4D parallelism in the Llama architecture.
Additionally, based on empirical insights from 454 experiments conducted on A100(40GB) and H100(94GB) GPUs, we provide analytical results that consider factors such as temporary buffers and fragmentation, which are difficult to theoretically calculate in terms of memory usage. From our experiments, we found that when the memory consumption estimated by our estimator is 80\% or less of the GPU memory, training succeeds in all cases.
This simple yet effective equation not only allows us to detect parallelism configurations that would result in out-of-memory beforehand and reduce the configuration search space, but also enables exploration without relying on empirically optimal configurations.
Furthermore, we analyzed the results obtained from 454 experiments to explore optimal settings in 4D parallelism, for which comprehensive configuration insights are lacking.

\section{Related work}

\subsection{Parallelism}
\label{sec:related-work:parallelism}

Model parallelism enables training large models across multiple GPUs. Model parameters, optimizer states, and activations require a huge amount of memory and do not fit on a single GPU.
Even if we are able to fit the model on a single GPU (e.g., by using CPU offloading~\cite{ren2021zerooffloaddemocratizingbillionscalemodel}), the high number of compute operations required can result in unrealistically long training times.


The 3D parallelism adopted in Megatron-LM~\cite{megatron-pipeline} employs tensor parallelism and pipeline parallelism to distribute model parameters and optimizer states, enabling training within GPU memory constraints.
Tensor parallelism splits and parallelizes each layer's parameters across multiple GPUs, while pipeline parallelism partitions the model along the layer dimension.
The integration of ZeRO Stage 1~\cite{zero} and Megatron-LM's distributed optimizer, which shard optimizer states across data-parallel ranks, further enhances memory efficiency and training scalability.

Despite its benefits, tensor parallelism in Megatron-LM does not partition activations from layers such as Dropout and LayerNorm, leading to redundant activation memory across tensor parallel ranks.
Sequence Parallelism~\cite{korthikanti2023reducing} addresses this limitation by partitioning these activations along the sequence dimension, significantly reducing activation memory without incurring additional computational or communication overhead.
By effectively combining sequence parallelism with tensor parallelism, it is possible to enhance memory efficiency when training large models. 
Consequently, modern LLM training often employs a combination of 3D parallelism, distributed optimizers, and sequence parallelism.

\paragraph{Context Parallelism}
\label{sec:related-work:context-parallel}


Context Parallelism~\cite{context_parallel} (CP) is a method that performs parallelization along the sequence length dimension.
Unlike Sequence Parallelism~\cite{korthikanti2023reducing}, which only parallelizes the activations of Dropout and LayerNorm, Context Parallelism enables partitioning of the model's network inputs and all activations along the sequence dimension.

Llama 2~\cite{touvron2023llama2openfoundation} had a sequence length of 4,096 tokens, but Llama 3~\cite{dubey2024llama3herdmodels} increased this to 8,192 tokens, and Llama 3.1 further extended it to 131,072 tokens. As efficient training that supports long contexts is increasingly demanded, context parallelism—which allows partitioning along the sequence dimension—is extremely useful for reducing the activations per GPU.

In components other than self-attention, there are no inter-token operations; thus, introducing context parallelism does not alter the operation. However, in self-attention layers, inter-token operations occur, necessitating the gathering of the full sequence, which requires additional all-gather communications between GPUs in the forward pass. During backpropagation, reduce-scatter is applied to the activation gradients, and each GPU stores only its sequence chunk to reduce the activation memory footprint.

To date, there is no comprehensive performance evaluation of 4D parallelism (DP, TP, PP, CP) utilizing context parallelism, and knowledge for applying it to actual LLM training is lacking.

\subsection{Model states' memory consumption}
\label{sec:related-work:model-states}
Model states—including parameters, gradients, and optimizer states (such as momentum and variance in the Adam optimizer) are the primary consumers of GPU memory during training. 
Efficient management of these states is critical to scaling up LLM training without exceeding memory constraints.

In FP16/FP32 mixed-precision training, if the number of model parameters is $\Psi$, the parameters and gradients are stored in FP16 format (2 bytes per value), consuming $2\Psi$ bytes each. 
Additionally, the optimizer states—including parameters, momentum, and variance—are stored in FP32 format (4 bytes per value), consuming $4\Psi$ bytes each. Therefore, the total estimated memory consumption is $16\Psi$ bytes~\cite{zero}.

With the advent of A100 GPUs that support BF16 training, FP16/FP32 mixed-precision training has transitioned to BF16/FP32 mixed-precision training. 
Furthermore, performing gradient accumulation in FP32 for numerical stability, as adopted in Llama-3~\cite{dubey2024llama3herdmodels} and implemented in widely used pre-training libraries like Megatron-LM\footnote{\url{https://github.com/NVIDIA/Megatron-LM}}, necessitates improvements to the formula presented in~\cite{zero} to be applicable for training the latest LLMs.

\subsection{Activation memory consumptions}
\label{sec:related-work:activation-memory}


The estimation of activations has been provided by~\cite{korthikanti2023reducing}. However, this formula is not only specific to the GPT architecture but also does not assume the use of FlashAttention~\cite{dao2022flashattention, dao2023flashattention}, which is utilized for faster and more memory-efficient training. Furthermore, it does not consider context parallelism~\cite{context_parallel}. As a result, it cannot accurately calculate the activation memory when training Llama architecture models using 4D Parallelism.

\section{Memory Usage}

\begin{figure*}[ht]
    \begin{center}
    \includegraphics[width=0.8\linewidth]{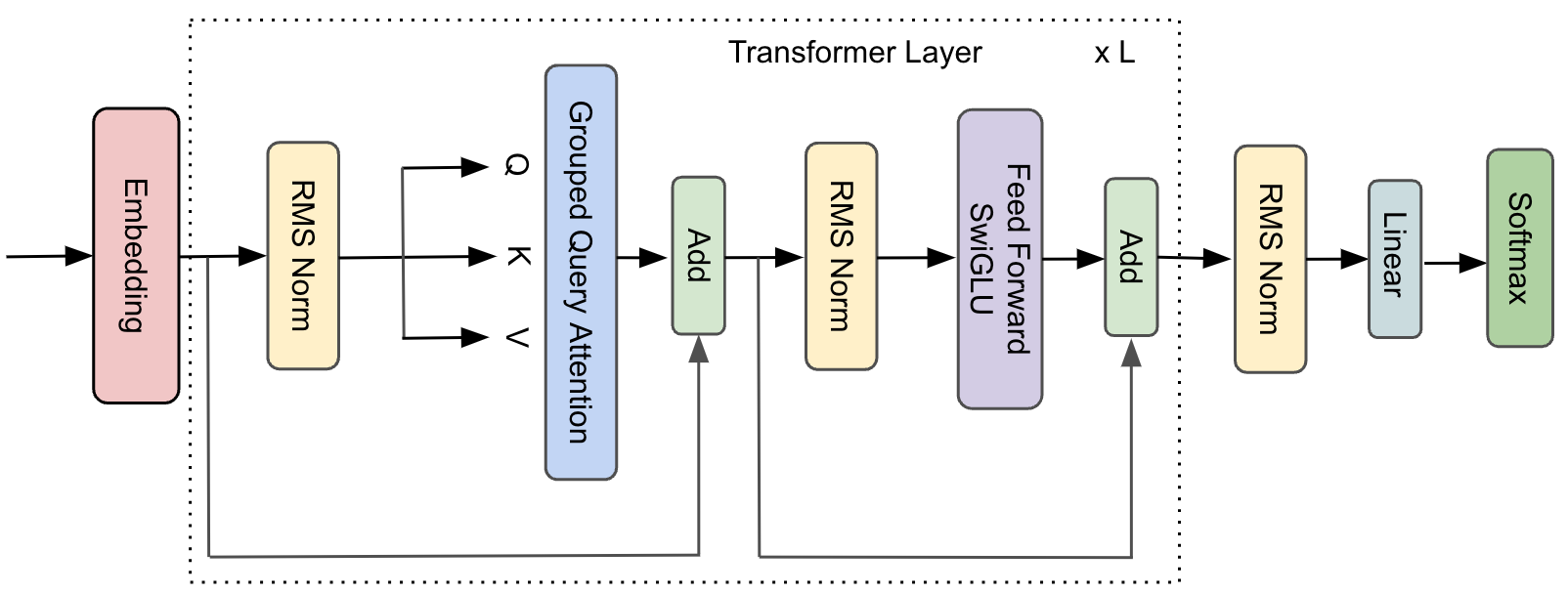}
    \end{center}
    \caption{Llama Architecture}
    \label{fig:llama-achitecture}
\end{figure*}


Previous studies~\cite{zero, korthikanti2023reducing} have attempted to consider GPU memory usage stemming from parameters, gradients, and optimizer states, as well as to calculate activation memory.
However, none have simultaneously considered both aspects to estimate and discuss the per-GPU memory requirements in the context of 4D parallelism.
Moreover, prior works have primarily focused on LLMs based on the GPT architecture and have not provided memory estimates that take into account the Llama architecture, which has been adopted by many open-source models.

In this study, we provide practical formulas to predict the memory consumption by estimating both parameter and activation memory using generalized equations that can handle the Llama architecture.

\subsection{Model states' memory}

\begin{table*}
    \centering
    \caption{Variable names.}
    \begin{tabular}{c|lc|l}
        $a$ & number of attention heads & $p$ & pipeline parallel size \\
        $b$ & microbatch size & $s$ & sequence length \\
        $h$ & hidden size & $t$ & tensor parallel size \\
        $d$ & data parallel size & $c$ & context parallel size \\
        $h_\text{ffn}$ & FFN hidden size & $k$ & number of key-value heads \\
        $L$ & number of transformer layers & $v$ & vocabulary size 
    \end{tabular}
    \label{tab:variables}
\end{table*}


We consider a single-stack transformer decoder model as shown in Figure~\ref{fig:llama-achitecture}.
The input tensor of size $b \times s \times h$ passes through an embedding layer of size $v \times h$, followed by $L$ transformer layers, and then through RMSNorm, a linear layer, and a Softmax function to produce the output.
Each transformer layer consists of self-attention, a feed-forward network (FFN), and RMSNorm. Notably, the Llama architecture does not include Dropout layers, which are present in the GPT architecture.
Additionally, in the transformer layer's FFN, the GPT architecture increases the hidden size to $4h$ and then reduces it back to $h$. To accommodate the Llama architecture, we introduce $h_{\text{ffn}}$ without assuming it to be $4h$, thereby defining a generalized variable. Furthermore, in the Llama architecture, the weights of the embedding layer and the language model head (output layer) are not shared; thus, we perform calculations under the assumption that they are unshared.

For reference, the variable names are listed in Table~\ref{tab:variables}. To estimate the memory size consumed by parameters and optimizer states, we first calculate the number of model parameters.

Let the weights of the self-attention be $Q = X W_Q$, $K = X W_K$, $V = X W_V$, and let $W_O$ be the weights of the linear layer after self-attention. 
Considering the group size in Grouped Query Attention as $g = \frac{a}{k}$, the sizes of $W_Q$, $W_K$, $W_V$, and $W_O$ are $(h, h)$, $(h, h/g)$, $(h, h/g)$, and $(h, h)$, respectively. Therefore, the number of attention parameters per layer is:

\begin{equation}
    \begin{split}
        \text{Attention parameter per layer} = 2h^2(1+\frac{k}{a} )
    \end{split}
    \label{eq:parameter-attention}
\end{equation}


Similarly, in the FFN layer, the sizes of the up projection $W_{\text{mlp\_up}}$, gate projection $W_{\text{mlp\_gate}}$, and down projection $W_{\text{mlp\_down}}$ are $(h, h_{\text{ffn}})$, $(h, h_{\text{ffn}})$, and $(h_{\text{ffn}}, h)$, respectively.
Therefore, the number of FFN parameters per layer is:

\begin{equation}
\text{FFN parameter per layer} = 3 h h_{ffn} 
\label{eq:parameter-ffn}
\end{equation}

Taking into account the embedding layer, language model head, RMSNorm layers, and the final RMSNorm, the total number of parameters can be expressed as:

\begin{equation}
\begin{split}
2hv + L(2h^2(1+\frac{k}{a} )+ 3h^2\frac{h_\text{ffn}}{h} + 2h) + h\\
=  2hv + h + 2Lh^2(1+\frac{k}{a} + \frac{3}{2}\frac{h_\text{ffn}}{h} + \frac{1}{h})
\end{split}
\label{eq:total-the-number-of-parameter}
\end{equation}

In Equation~\eqref{eq:total-the-number-of-parameter}, the term $2hv$ accounts for the embedding and output layers, $L$ is the number of transformer layers, and $h$ is the hidden size. 
The factor $2h^2\left(1+\frac{k}{a}\right)$ comes from the attention parameters per layer (Equation~\eqref{eq:parameter-attention}), and $3h h_{\text{ffn}}$ is from the FFN parameters per layer (Equation~\eqref{eq:parameter-ffn}). 
The additional $2h$ within the parentheses accounts for the RMSNorm layers in each transformer layer, and the final $h$ outside accounts for the last RMSNorm layer after the transformer stack.



From the above calculations, we have determined the number of parameters. Based on this, we can calculate the memory consumed by the parameter weights and optimizer states.
In ZeRO~\cite{zero}, since gradients are assumed to be stored in FP16 or BF16, the memory size required during training when using the Adam~\cite{kingma2017adammethodstochasticoptimization} optimizer is considered to be $16\Psi$, where $\Psi$ is the number of parameters. 
However, as reported by Llama 3~\cite{dubey2024llama3herdmodels}, gradients may be accumulated in FP32 to ensure convergence during training. In this paper, we assume that gradients are accumulated in FP32, resulting in a memory consumption of $18\Psi$.
Note that FP16/BF16 consumes 2 bytes per value, while FP32 consumes 4 bytes per value. The breakdown of the coefficient $18$ is as follows:

\begin{tabular}{@{}ll@{}}
  weight: & \textbf{BF16}(2 bytes) \\
  gradients: & \textbf{FP32}(4 bytes) \\
  optimizer states: & \\
  \quad parameter (master weight): & \textbf{FP32}(4 bytes) \\
  \quad gradient momentum: & \textbf{FP32}(4 bytes) \\
  \quad gradient variance: & \textbf{FP32}(4 bytes) \\
\end{tabular}

Therefore, the memory consumed by the parameters, gradients, and optimizer states is given by:

\begin{equation}
\begin{split}
18(2hv + h + 2Lh^2(1+\frac{k}{a} + \frac{3}{2}\frac{h_\text{ffn}}{h} + \frac{1}{h}))
\end{split}
\end{equation}

This equation holds in cases where model parallelism methods, such as Tensor Parallelism and Pipeline Parallelism, are not applied. 
In the following sections, we will sequentially demonstrate how the per-GPU memory consumption changes when each parallelization method is introduced.

\subsubsection{Data Parallelism}
\label{sec:model-states-data-parallel}
By utilizing techniques such as ZeRO Stage 1~\cite{zero} and Megatron-LM's distributed optimizer, which shard the optimizer states across data parallel processes, the optimizer states amounting to $12\Psi$ are distributed among the data parallel processes. As a result, the memory consumption per GPU becomes:

\begin{equation}
    \begin{split}
        (6 + \frac{12}{d})(2hv + h + 2Lh^2(1+\frac{k}{a} + \frac{3}{2}\frac{h_\text{ffn}}{h} + \frac{1}{h}))
    \end{split}
    \label{eq:memory-consumption-model-states-data-parallel}
\end{equation}

\subsubsection{Tensor Parallelism}
\label{sec:model-staes-tensor-parallel}
Tensor Parallelism divides the parameters of the Attention, FFN, embedding, and language model head layers by $t$. Specifically, the Attention parameters become $(2h^{2}(1+\frac{k}{a}))/t$, and the MLP parameters become $(3 h h_{\text{ffn}})/t$. 
Therefore, when both Data Parallelism and Tensor Parallelism are applied, the memory consumption of parameters, gradients, and optimizer states per GPU is:

\begin{equation}
    \begin{split}
        (6 + \frac{12}{d})(\frac{2hv}{t} + h + 2Lh^2(\frac{1+\frac{k}{a} + \frac{3}{2}\frac{h_\text{ffn}}{h}}{t} + \frac{1}{h}))
    \end{split}
    \label{eq:memory-consumption-model-states-tensor-parallel}
\end{equation}

\subsubsection{Pipeline Parallelism}
\label{sec:model-states-pipeline-parallel}

In Pipeline Parallelism, for the first pipeline stage—which includes the embedding layer—the memory consumption is calculated as follows. 
Since the final LayerNorm does not pertain to the first pipeline stage, only the embedding layer's parameters $hv$ and the parameters of $\frac{L}{p}$ layers are considered. 
In this paper, we assume the 1F1B pipeline schedule developed in PipeDream~\cite{1f1b-pipedream} for pipeline scheduling.

\begin{equation}
    \begin{split}
        (6 + \frac{12}{d})(\frac{hv}{t} + 2\frac{L}{p}h^2(\frac{1+\frac{k}{a} + \frac{3}{2}\frac{h_\text{ffn}}{h}}{t} + \frac{1}{h}))
    \end{split}
    \label{eq:memory-consumption-model-states-pipeline-parallel}
\end{equation}

For the intermediate pipeline stages, which do not include the embedding or language model head layer, the memory consumption per GPU is:

\begin{equation}
\begin{split}
(6 + \frac{12}{d})(2\frac{L}{p}h^2(\frac{1+\frac{k}{a} + \frac{3}{2}\frac{h_\text{ffn}}{h}}{t} + \frac{1}{h}))
\end{split}
\end{equation}

For the last pipeline stage, considering the language model head and the final LayerNorm (RMSNorm), the memory consumption per GPU is:

\begin{equation}
\begin{split}
(6 + \frac{12}{d})(\frac{hv}{t} + h + 2\frac{L}{p}h^2(\frac{1+\frac{k}{a} + \frac{3}{2}\frac{h_\text{ffn}}{h}}{t} + \frac{1}{h}))
\end{split}
\end{equation}

\subsubsection{Context Parallelism}
\label{sec:model-states-context-parallel}
Context Parallelism not only partitions the inputs and activations along the sequence dimension but also distributes the optimizer states across processes. 
Consequently, the optimizer states are divided among $d \times c$ processes. Considering this, the memory consumption per GPU (here shown for the first stage of Pipeline Parallelism) is:

\begin{equation}
\begin{split}
(6 + \frac{12}{d c})(\frac{hv}{t} + 2\frac{L}{p}h^2(\frac{1+\frac{k}{a} + \frac{3}{2}\frac{h_\text{ffn}}{h}}{t} + \frac{1}{h}))
\end{split}
\label{eq:model-states-memory-per-gpu}
\end{equation}

\subsection{Activation Memory}

\begin{figure}[ht]
    \begin{center}
    \includegraphics[width=1.0\linewidth]{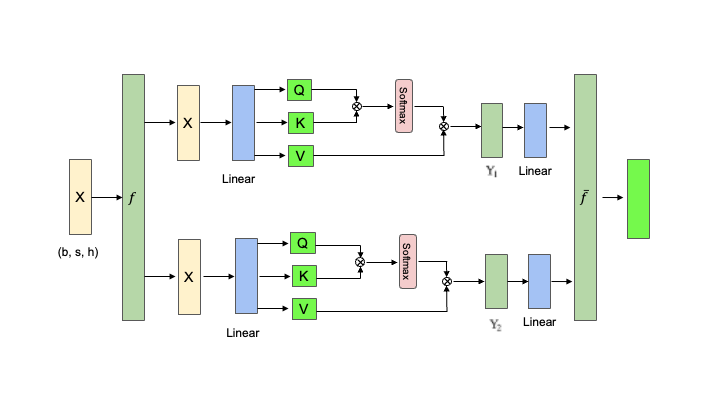}
    \end{center}
    \caption{Attention Block (when tensor parallel size = 2)}
    \label{fig:attention}
\end{figure}

\begin{figure}[ht]
    \begin{center}
    \includegraphics[width=0.8\linewidth]{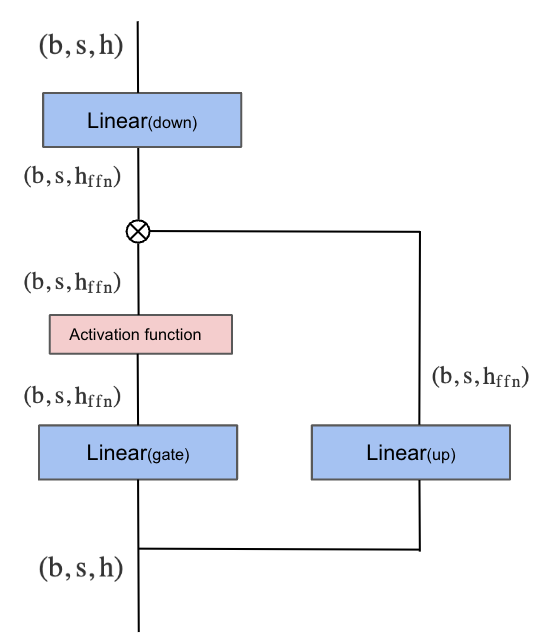}
    \end{center}
    \caption{FFN (Feed Forward Network) in the Llama architecture.}
    \label{fig:swiglu}
\end{figure}


In the previous section, we calculated the memory size consumed by parameters, gradients, and optimizer states. In this section, we compute the memory size consumed by activations.
As shown in Figure~\ref{fig:llama-achitecture}, the primary contributor to activation memory is the Transformer layer. Therefore, we first calculate the activations for the Attention block, Feed-Forward Network (FFN), and RMSNorm within the Transformer layer, and then consider the embedding layer and the language model (LM) head.
As depicted in Figure~\ref{fig:attention}, the attention block consists of self-attention followed by a linear projection.\footnote{While the GPT architecture includes Dropout layers, the Llama architecture does not include Dropout.} 
In practical LLM training, FlashAttention 2~\cite{dao2023flashattention} is used to improve training speed and reduce memory consumption. Therefore, we perform the activation calculations assuming that FlashAttention 2 is being utilized.


\begin{itemize} 
    \item \textbf{Query ($Q$), Key ($K$), and Value ($V$) matrix multiplies:} The size of the Query is $(b, s, h)$, so the required activations are $2sbh$. However, when employing Grouped Query Attention~\cite{ainslie2023gqatraininggeneralizedmultiquery}, the Key and Value tensors have sizes of $(b, s, h \times \frac{k}{a})$, so the activations required for each are $2sbh \times \frac{k}{a}$. 
    \item \textbf{$QK^\top$ matrix multiply:} FlashAttention does not store the results of the large $QK^\top$ matrices and recomputes them during the backward pass. Therefore, the activation memory required is zero. 
    \item \textbf{Softmax:} Since FlashAttention recomputes during the backward pass, the activation memory required for the Softmax is also zero. 
    \item \textbf{Attention over Values ($V$):} The output tensor size is $(b, s, h)$, so it consumes $2sbh$ of memory. 
\end{itemize}

Including the input $X$, the total activation memory consumed by the self-attention is:

\begin{equation}
    6sbh + 4sbh \times \frac{k}{a}
    \label{eq:activation-attention}
\end{equation}

\textbf{FFN:} The structure of the Llama architecture's FFN is shown in Figure~\ref{fig:swiglu}. Therefore, we require activations for backpropagation as follows: $2sbh$ for the input, $2sbh_{\text{ffn}}$ for the output of the up projection, $2sbh_{\text{ffn}}$ for the output of the gate projection, and $2sbh_{\text{ffn}}$ for the activation function non-linearity. Additionally, $2sbh_{\text{ffn}}$ is needed for the input to the down projection. In total, $2sb(h + 4h_{\text{ffn}})$ activations are required.



\textbf{Layer Norm:} Each RMSNorm stores its input with size $2sbh$, so in total, we need $4sbh$ of storage.
Adding up the memory required for the attention, FFN, and the RMSNorms, the total activation memory required to store the activations for a single layer of the Transformer network is:

\begin{equation} 
    \text{Activation memory per layer} = sbh \left(12 + 4\frac{k}{a} + 8\frac{h_{\text{ffn}}}{h}\right). 
    \label{eq:activation-single-transformer-layer} 
\end{equation}

The above equation applies to the case where no form of model parallelism is applied.

Next, we consider the activation memory consumed by the embedding layer and the language model head.

\textbf{Embedding:} The input to the embedding layer has a numerical precision of \texttt{int64}, which consumes 8 bytes per value. To keep the pipeline fully utilized and avoid idle time, the first stage must store activations for $p$ microbatches (for more details, see Figure 4 top of \cite{megatron-pipeline}). Therefore, the activation memory consumed by the embedding layer is:
\begin{equation}
    8 s b h p
\end{equation}

\textbf{LM Head:} The cross-entropy loss requires FP32 numerical precision, consuming $4sbv$ bytes of memory. Additionally, $2sbh$ is needed for the output RMSNorm, and $2sbh$ for the output linear layer, resulting in a total memory consumption of:
\begin{equation} 
    4sbh \left(1 + \frac{v}{h}\right)
\end{equation}
For more details, see Section 4.3 of \cite{korthikanti2023reducing}.

\subsubsection{Tensor Parallelism}



We use the tensor parallelism developed by \cite{megatron} to parallelize the attention and FFN modules, as illustrated in Figure~\ref{fig:attention}. 
This form of parallelism introduces two additional communication operations, $f$ and $\bar{f}$, per layer. For more details, please refer to \cite{megatron}.

As described in Section~\ref{sec:model-staes-tensor-parallel}, tensor parallelism parallelizes the model states, and similarly, the activations are also parallelized. 
However, certain elements, such as the input to the FFN, cannot be parallelized. Additionally, RMSNorm is not parallelized. 
By enabling Sequence Parallelism, we can parallelize these components as well. For more details, please see Figure 6 in \cite{korthikanti2023reducing}.

Considering the above, the activation memory size consumed by the Transformer block can be expressed using the tensor parallel size $t$ as follows:

\begin{equation}
    \text{Transformer layers = } sbh \left(12 + 4\frac{k}{a} + 8\frac{h_\text{ffn}}{h}\right) L / t
    \label{eq:activation-memory-tensor-parallel}
\end{equation}

\subsubsection{Pipeline Parallelism}

\begin{figure}[ht]
    \begin{center}
    \includegraphics[width=1.0\linewidth]{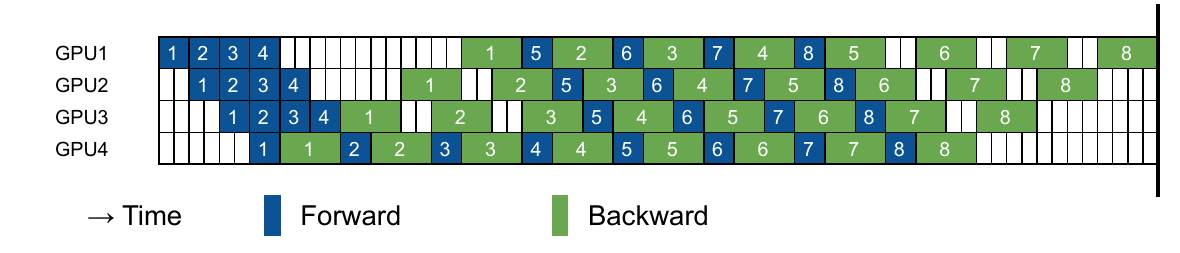}
    \end{center}
    \caption{1F1B (One Forward One Backward) Pipeline schedule. Blue represents the forward pass, green indicates the backward pass, and uncolored spaces represent pipeline bubbles.}
    \label{fig:1f1b-pipeline-schedule}
\end{figure}


Pipeline Parallelism divides the Transformer layers, which consist of $L$ layers, into $p$ groups, each containing $L/p$ layers. 
However, unlike in Section~\ref{sec:model-states-pipeline-parallel}, the total activation consumed by the Transformer layers is not evenly divided by $p$. 
This is due to the pipeline parallel scheduling. 
In this paper, we assume the 1F1B pipeline schedule, as shown in Figure~\ref{fig:1f1b-pipeline-schedule}. 
In the 1F1B schedule, the GPU assigned to the first pipeline stage has up to $p$ microbatches worth of activations, while the GPU assigned to the last pipeline stage needs to store activations for only one microbatch because the backward pass starts immediately. 
Therefore, although each stage is responsible for $L/p$ layers, the first stage holds activations equivalent to $L/p \times p = L$ layers.
Since this does not depend on $p$, the first stage always needs to store the same amount of activations for $L$ layers, regardless of the pipeline parallel size $p$.

Considering the above and introducing the Kronecker delta $\delta_{p,1}$, the maximum activation memory per GPU when using Pipeline Parallelism can be expressed as follows. 
Here, we focus only on the first pipeline stage, which is critical in determining whether an out-of-memory (OOM) occurs.

\begin{equation}
   \frac{sbh}{t} \left((12 + 4\frac{k}{a} + 8\frac{h_\text{ffn}}{h}) L + 8p + \delta_{p, 1} 4(1+v/h)\right)
\end{equation}

\subsubsection{Context Parallelism}


As explained in Section~\ref{sec:related-work:context-parallel}, context parallelism partitions the network inputs and all activations along the sequence dimension. 
Therefore, the activation memory consumption per GPU can be expressed using the context parallel size $c$ as:

\begin{equation}
   \frac{sbh}{tc} \left((12 + 4\frac{k}{a} + 8\frac{h_\text{ffn}}{h}) L + 8p + \delta_{p, 1} 4(1+v/h)\right)
   \label{eq:activation-memory-per-gpu}
\end{equation}

\subsection{Total Memory}
\label{sec:activation-total-memory-consumption}

The majority of the required activation memory per GPU is captured by Equation~\ref{eq:activation-memory-per-gpu}.
Similarly, the required memory per GPU for model states (parameters, gradients, optimizer states) is calculated using Equation~\ref{eq:model-states-memory-per-gpu}.
Therefore, excluding residual memory consumption due to temporary buffers and memory fragmentation—which are difficult to calculate theoretically—the total memory consumption is given by:

\begin{multline}
   \left(6 + \frac{12}{d c}\right)\left(\frac{hv}{t} + 2\frac{L}{p}h^2\left(\frac{1+\frac{k}{a} + \frac{3}{2}\frac{h_\text{ffn}}{h}}{t} + \frac{1}{h}\right)\right) \\
   + \frac{sbh}{tc} \left((12 + 4\frac{k}{a} + 8\frac{h_\text{ffn}}{h}) L + 8p + \delta_{p, 1} \, 4\left(1+\frac{v}{h}\right)\right)
   \label{eq:total-memory-per-gpu}
\end{multline}

\section{Evaluations}

\subsection{Memory Usage}
\label{sec:eval:memory-usage}

\begin{table*}[ht]
    \centering
    \caption{Estimated Total Memory (GB) per GPU when training Llama-3.1-8B on A100(40GB) with sequence length 8,192. The vertical axis (TP, CP, PP, MBS) represents Tensor Parallel size (TP), Context Parallel size (CP), Pipeline Parallel size (PP), and Micro Batch Size (MBS), respectively. Red cells indicate settings that are estimated to exceed the 40GB GPU memory limit, likely resulting in Out of Memory (OOM) errors. Yellow cells denote configurations that may approach the limit due to temporary buffers and memory fragmentation, and green cells represent estimations at or below 80\% of GPU memory (32GB or less), suggesting they are less likely to cause OOM.}
    \begin{tabular}{|c|c|c|c|c|c|c|}
        \hline
        (TP, CP, PP, MBS) & 8 GPUs & 16 GPUs & 32 GPUs & 64 GPUs & 128 GPUs & 256 GPUs \\
        \hline
        (4, 1, 2, 1) & \cellcolor{lightgreen} 27.2 & \cellcolor{lightgreen}21.59 & \cellcolor{lightgreen}18.79 & \cellcolor{lightgreen}17.39 & \cellcolor{lightgreen}16.69 & \cellcolor{lightgreen}16.34 \\
        (4, 1, 2, 2) & \cellcolor{lightyellow}37.58 & \cellcolor{lightgreen} 31.97 & \cellcolor{lightgreen}29.16 & \cellcolor{lightgreen}27.76 & \cellcolor{lightgreen}27.06 & \cellcolor{lightgreen}26.71 \\
        (4, 1, 2, 4) & \cellcolor{lightred}58.33 & \cellcolor{lightred}52.72 & \cellcolor{lightred}49.91 & \cellcolor{lightred}48.51 & \cellcolor{lightred}47.81 & \cellcolor{lightred}47.46 \\ \hline
        (4, 2, 2, 1) & - & \cellcolor{lightgreen}16.41 & \cellcolor{lightgreen}13.6 & \cellcolor{lightgreen}12.2 & \cellcolor{lightgreen}11.5 & \cellcolor{lightgreen}11.15 \\
        (4, 2, 2, 2) & - & \cellcolor{lightgreen}21.59 & \cellcolor{lightgreen}18.79 & \cellcolor{lightgreen}17.39 & \cellcolor{lightgreen}16.69 & \cellcolor{lightgreen}16.34 \\
        (4, 2, 2, 4) & -  & \cellcolor{lightgreen} 31.97 & \cellcolor{lightgreen}29.16 & \cellcolor{lightgreen}27.76 & \cellcolor{lightgreen}27.06 & \cellcolor{lightgreen}26.71 \\
        (4, 2, 2, 8) & - & \cellcolor{lightred}52.72 & \cellcolor{lightred}49.91 & \cellcolor{lightred}48.51 & \cellcolor{lightred}47.81 & \cellcolor{lightred}47.46 \\ \hline
        (2, 2, 2, 1) & \cellcolor{lightyellow}32.81 & \cellcolor{lightgreen}27.2 & \cellcolor{lightgreen}24.4 & \cellcolor{lightgreen}23.00 & \cellcolor{lightgreen}22.29 & \cellcolor{lightgreen}21.94 \\
        (2, 2, 2, 2) & \cellcolor{lightred}43.19 & \cellcolor{lightyellow}37.58 & \cellcolor{lightyellow}34.77 & \cellcolor{lightyellow}33.37 & \cellcolor{lightyellow}32.67 & \cellcolor{lightyellow}32.32 \\ \hline
        (2, 4, 2, 1) & -  & \cellcolor{lightgreen}22.02 & \cellcolor{lightgreen}19.21 & \cellcolor{lightgreen}17.81 & \cellcolor{lightgreen}17.11 & \cellcolor{lightgreen}16.76 \\
        (2, 4, 2, 2) & - & \cellcolor{lightgreen}27.2 & \cellcolor{lightgreen}24.40 & \cellcolor{lightgreen}23.00 & \cellcolor{lightgreen}22.29 & \cellcolor{lightgreen}21.94 \\ \hline
        (4, 2, 1, 1) & \cellcolor{lightgreen}28.1 & \cellcolor{lightgreen}22.49 & \cellcolor{lightgreen}19.69 & \cellcolor{lightgreen}18.28 & \cellcolor{lightgreen}17.58 & \cellcolor{lightgreen}17.23 \\
        (4, 2, 1, 2) & \cellcolor{lightyellow}33.76 & \cellcolor{lightgreen}28.15 & \cellcolor{lightgreen}25.35 & \cellcolor{lightgreen}23.94 & \cellcolor{lightgreen}23.24 & \cellcolor{lightgreen}22.89 \\
        (4, 2, 1, 4) & \cellcolor{lightred}45.08 & \cellcolor{lightyellow}39.47 & \cellcolor{lightyellow}36.67 & \cellcolor{lightyellow}35.27 & \cellcolor{lightyellow}34.56 & \cellcolor{lightyellow}34.21 \\ \hline
        (2, 2, 4, 1) & - & \cellcolor{lightgreen}23.19 & \cellcolor{lightgreen}20.01 & \cellcolor{lightgreen}18.43 & \cellcolor{lightgreen}17.64 & \cellcolor{lightgreen}17.24 \\
        (2, 2, 4, 2) & - & \cellcolor{lightyellow}33.69 & \cellcolor{lightgreen}30.51 & \cellcolor{lightgreen}28.93 & \cellcolor{lightgreen}28.14 & \cellcolor{lightgreen}27.74 \\
        (2, 2, 4, 4) & - & \cellcolor{lightred}54.69 & \cellcolor{lightred}51.51 & \cellcolor{lightred}49.93 & \cellcolor{lightred}49.14 & \cellcolor{lightred}48.74 \\ \hline
        (2, 4, 1, 1) & \cellcolor{lightyellow}39.32 & \cellcolor{lightyellow}33.71 & \cellcolor{lightgreen}30.9 & \cellcolor{lightgreen}29.5 & \cellcolor{lightgreen}28.8 & \cellcolor{lightgreen}28.45 \\
        (2, 4, 1, 2) & \cellcolor{lightred}44.98 & \cellcolor{lightyellow}39.37 & \cellcolor{lightyellow}36.56 & \cellcolor{lightyellow}35.16 & \cellcolor{lightyellow}34.46 & \cellcolor{lightyellow}34.11 \\
        (2, 4, 1, 4) & \cellcolor{lightred}56.3 & \cellcolor{lightred}50.69 & \cellcolor{lightred}47.89 & \cellcolor{lightred}46.48 & \cellcolor{lightred}45.78 & \cellcolor{lightred}45.43 \\ \hline
        (4, 1, 1, 1) & \cellcolor{lightyellow}33.76 & \cellcolor{lightgreen}28.15 & \cellcolor{lightgreen}25.35 & \cellcolor{lightgreen}23.94 & \cellcolor{lightgreen}23.24 & \cellcolor{lightgreen}22.89 \\
        (4, 1, 1, 2) & \cellcolor{lightred}45.08 & \cellcolor{lightyellow}39.47 & \cellcolor{lightyellow}36.67 & \cellcolor{lightyellow}35.27 & \cellcolor{lightyellow}34.56 & \cellcolor{lightyellow}34.21 \\ \hline
        (2, 2, 1, 1) & \cellcolor{lightred}44.98 & \cellcolor{lightyellow}39.37 & \cellcolor{lightyellow}36.56 & \cellcolor{lightyellow}35.16 & \cellcolor{lightyellow}34.46 & \cellcolor{lightyellow}34.11 \\
        (2, 2, 1, 2) & \cellcolor{lightred}56.3 & \cellcolor{lightred}50.69 & \cellcolor{lightred}47.89 & \cellcolor{lightred}46.48 & \cellcolor{lightred}45.78 & \cellcolor{lightred}45.43 \\ \hline
        (2, 1, 2, 1) & \cellcolor{lightred}43.19 & \cellcolor{lightyellow}37.58 & \cellcolor{lightyellow}34.77 & \cellcolor{lightyellow}33.37 & \cellcolor{lightyellow}32.67 & \cellcolor{lightyellow}32.32 \\
        (2, 1, 2, 2) & \cellcolor{lightred}63.94 & \cellcolor{lightred}58.33 & \cellcolor{lightred}55.52 & \cellcolor{lightred}54.12 & \cellcolor{lightred}53.42 & \cellcolor{lightred}53.07 \\
        \hline
    \end{tabular}
    \label{tab:a100-memory-table}
\end{table*}

\begin{table*}[h]
    \centering
    \caption{Measured throughput(TFLOP/s) when training Llama-3.1-8B on A100 (40GB) with a sequence length of 8,192. The colors red, yellow, and green correspond to memory consumption levels predicted by the memory consumption estimator: red indicates configurations that likely exceed memory limits, yellow denotes configurations near the memory limit, and green represents configurations predicted to use 80\% or less of GPU memory (32GB or less).}
    \begin{tabular}{|c|c|c|c|c|c|c|}
        \hline
        (TP, CP, PP, MBS) & 8 GPUs & 16 GPUs & 32 GPUs & 64 GPUs & 128 GPUs & 256 GPUs \\
        \hline
        (4, 1, 2, 1) & \cellcolor{lightgreen}182.6 & \cellcolor{lightgreen}178.38 & \cellcolor{lightgreen}176.79 & \cellcolor{lightgreen}170.84 & \cellcolor{lightgreen}169.52 & \cellcolor{lightgreen}153.9 \\
        (4, 1, 2, 2) & \cellcolor{lightyellow}OOM & \cellcolor{lightgreen}189.79 & \cellcolor{lightgreen}187.01 & \cellcolor{lightgreen}185.69 & \cellcolor{lightgreen}182.28 & \cellcolor{lightgreen}178.62 \\
        (4, 1, 2, 4) & \cellcolor{lightred}OOM & \cellcolor{lightred}OOM & \cellcolor{lightred}OOM & \cellcolor{lightred}OOM & \cellcolor{lightred}OOM & \cellcolor{lightred}OOM \\ \hline
        (4, 2, 2, 1) & - & \cellcolor{lightgreen}151.65 & \cellcolor{lightgreen}148.7 & \cellcolor{lightgreen}142.34 & \cellcolor{lightgreen}134.34 & \cellcolor{lightgreen}116.28 \\
        (4, 2, 2, 2) & - & \cellcolor{lightgreen}155.01 & \cellcolor{lightgreen}160.27 & \cellcolor{lightgreen}158.15 & \cellcolor{lightgreen}150.64 & \cellcolor{lightgreen}127.03 \\
        (4, 2, 2, 4) & - & \cellcolor{lightgreen}158.99 & \cellcolor{lightgreen}156.36 & \cellcolor{lightgreen}154.88 & \cellcolor{lightgreen}160.62 & \cellcolor{lightgreen}160.83 \\
        (4, 2, 2, 8) & - & \cellcolor{lightred}OOM & \cellcolor{lightred}OOM & \cellcolor{lightred}OOM & \cellcolor{lightred}OOM & \cellcolor{lightred}OOM \\ \hline
        (2, 2, 2, 1) & \cellcolor{lightyellow}187.83 & \cellcolor{lightgreen}192.62 & \cellcolor{lightgreen}185.07 & \cellcolor{lightgreen}178.58 & \cellcolor{lightgreen}165.2 & \cellcolor{lightgreen}142.61 \\
        (2, 2, 2, 2) & \cellcolor{lightred}OOM & \cellcolor{lightyellow}OOM & \cellcolor{lightyellow}196.65 & \cellcolor{lightyellow}192.54 & \cellcolor{lightyellow}188.77 & \cellcolor{lightyellow}183.17 \\ \hline
        (2, 4, 2, 1) & - & \cellcolor{lightgreen}133.54 & \cellcolor{lightgreen}128.95 & \cellcolor{lightgreen}124.56 & \cellcolor{lightgreen}108.73 & \cellcolor{lightgreen}98.79 \\
        (2, 4, 2, 2) & - & \cellcolor{lightgreen}144.41 & \cellcolor{lightgreen}142.4 & \cellcolor{lightgreen}138.89 & \cellcolor{lightgreen}124.05 & \cellcolor{lightgreen}117.65 \\ \hline
        (4, 2, 1, 1) & \cellcolor{lightgreen}171.16 & \cellcolor{lightgreen}174.07 & \cellcolor{lightgreen}171.99 & \cellcolor{lightgreen}168.66 & \cellcolor{lightgreen}162.15 & \cellcolor{lightgreen}153.03 \\
        (4, 2, 1, 2) & \cellcolor{lightyellow}OOM & \cellcolor{lightgreen}188.97 & \cellcolor{lightgreen}187.29 & \cellcolor{lightgreen}185.35 & \cellcolor{lightgreen}183.49 & \cellcolor{lightgreen}181.97 \\
        (4, 2, 1, 4) & \cellcolor{lightred}OOM & \cellcolor{lightyellow}OOM & \cellcolor{lightyellow}OOM & \cellcolor{lightyellow}OOM & \cellcolor{lightyellow}OOM & \cellcolor{lightyellow}OOM \\ \hline
        (2, 2, 4, 1) & - & \cellcolor{lightgreen}153.99 & \cellcolor{lightgreen}137.78 & \cellcolor{lightgreen}133.72 & \cellcolor{lightgreen}121.44 & \cellcolor{lightgreen}105.93 \\
        (2, 2, 4, 2) & - & \cellcolor{lightyellow}157.73 & \cellcolor{lightgreen}143.15 & \cellcolor{lightgreen}135.33 & \cellcolor{lightgreen}126.91 & \cellcolor{lightgreen}118.14 \\
        (2, 2, 4, 4) & - & \cellcolor{lightred}OOM & \cellcolor{lightred}OOM & \cellcolor{lightred}OOM & \cellcolor{lightred}OOM & \cellcolor{lightred}OOM \\ \hline
        (2, 4, 1, 1) & \cellcolor{lightyellow}OOM & \cellcolor{lightyellow}OOM & \cellcolor{lightgreen}163.1 & \cellcolor{lightgreen}153.42 & \cellcolor{lightgreen}144.44 & \cellcolor{lightgreen}131.53 \\
        (2, 4, 1, 2) & \cellcolor{lightred}OOM & \cellcolor{lightyellow}OOM & \cellcolor{lightyellow}OOM & \cellcolor{lightyellow}OOM & \cellcolor{lightyellow}OOM & \cellcolor{lightyellow}OOM \\
        (2, 4, 1, 4) & \cellcolor{lightred}OOM & \cellcolor{lightred}OOM & \cellcolor{lightred}OOM & \cellcolor{lightred}OOM & \cellcolor{lightred}OOM & \cellcolor{lightred}OOM \\ \hline
        (4, 1, 1, 1) & \cellcolor{lightyellow}\textbf{194.25} & \cellcolor{lightgreen}\textbf{194.97} & \cellcolor{lightgreen}194.64 & \cellcolor{lightgreen}190.9 & \cellcolor{lightgreen}191.88 & \cellcolor{lightgreen}188.28 \\
        (4, 1, 1, 2) & \cellcolor{lightred}OOM & \cellcolor{lightyellow}OOM & \cellcolor{lightyellow}OOM & \cellcolor{lightyellow}OOM & \cellcolor{lightyellow}\textbf{201.37} & \cellcolor{lightyellow}\textbf{197.12} \\ \hline
        (2, 2, 1, 1) & \cellcolor{lightred}OOM & \cellcolor{lightyellow}OOM & \cellcolor{lightyellow}OOM & \cellcolor{lightyellow}OOM & \cellcolor{lightyellow}198.59 & \cellcolor{lightyellow}193.57 \\
        (2, 2, 1, 2) & \cellcolor{lightred}OOM & \cellcolor{lightred}OOM & \cellcolor{lightred}OOM & \cellcolor{lightred}OOM & \cellcolor{lightred}OOM & \cellcolor{lightred}OOM \\ \hline
        (2, 1, 2, 1) & \cellcolor{lightred}OOM & \cellcolor{lightyellow}OOM & \cellcolor{lightyellow}\textbf{203.98} & \cellcolor{lightyellow}\textbf{202.00} & \cellcolor{lightyellow}195.55 & \cellcolor{lightyellow}193.48 \\
        (2, 1, 2, 2) & \cellcolor{lightred}OOM & \cellcolor{lightred}OOM & \cellcolor{lightred}OOM & \cellcolor{lightred}OOM & \cellcolor{lightred}OOM & \cellcolor{lightred}OOM \\
        \hline
    \end{tabular}
    \label{tab:a100-tflops}
\end{table*}

We utilized the memory consumption estimator implemented based on the equations presented in this paper to conduct experiments on NVIDIA A100 SXM (40GB) and NVIDIA H100 SXM (94GB) GPUs.
We conducted experiments training models with the same architecture as Llama-3.1-8B. On the A100, we used a sequence length of 8,192, while on the H100, we experimented with sequence lengths of 8,192, 16,384, and 32,768.
Additionally, we trained models with the same architecture as Llama-3.1-70B on the A100 with a sequence length of 8,192 only.
In all experiments, the global batch size was fixed at 1,024.

The memory consumption estimator-predicted values represent the memory consumption of model states and activations, as described in Section~\ref{sec:activation-total-memory-consumption}, and do not account for temporary buffers and fragmentation.
From our experimental results, we found that if there is 20\% spare GPU memory, the impact of temporary buffers and fragmentation does not affect whether an out-of-memory (OOM) occurs, as demonstrated in Table~\ref{tab:a100-tflops} by actually experimenting with each configuration.
This finding was consistent not only in the A100 (40GB) environment but also in the H100 (94GB) environment, as shown in Table~\ref{tab:h100-tflops-8192}, \ref{tab:h100-tflops-8B-16384}, \ref{tab:h100-tflops-8B-32768}.

Therefore, by using Equation~\ref{eq:total-memory-per-gpu}, we demonstrated that it is possible to estimate, without the need for an accelerator device environment, which configurations are trainable within the memory of the devices used in any GPU environment, which configurations are not, and which configurations may be trainable depending on temporary buffers and fragmentation.
By using this simple yet effective equation, we could preliminarily narrow down candidate parallelism configurations that have the potential to achieve high throughput (TFLOP/s) by effectively utilizing HBM memory. 
By experimenting only with these candidates, it became possible to discover the optimal parallelism configuration while saving computational resources.

\subsection{Peformance Analysis of Parallelism Configuration}

In this section, we provide an empirical analysis of the results obtained from experiments using 4D parallelism on NVIDIA A100 SXM (40GB) and NVIDIA H100 SXM (94GB) GPUs.
Throughput was measured in TFLOP/s, using the formula adapted for the Llama architecture based on prior work~\cite{megatron-pipeline}.

\begin{table*}[h]
\centering
\caption{Measured throughput(TFLOP/s) when training Llama-3.1-8B on H100 (94GB) with a sequence length of 8,192. The colors red, yellow, and green correspond to memory consumption levels predicted by the memory consumption estimator: red indicates configurations that likely exceed memory limits, yellow denotes configurations near the memory limit, and green represents configurations predicted to use 80\% or less of GPU memory (75.2GB or less).}
\begin{tabular}{|c|c|c|c|c|c|c|c|}
\hline
(TP, CP, PP, MBS) & 4 GPUs & 8 GPUs & 16 GPUs & 32 GPUs & 64 GPUs \\ \hline
(2, 1, 1, 1) & \cellcolor{lightgreen}446.72 & \cellcolor{lightgreen}443.45 & \cellcolor{lightgreen}439.87 & \cellcolor{lightgreen}438.69 & \cellcolor{lightgreen}435.34 \\
(2, 1, 1, 2) & \cellcolor{lightyellow}OOM & \cellcolor{lightyellow}479.16 & \cellcolor{lightgreen}475.2 & \cellcolor{lightgreen}471.82 & \cellcolor{lightgreen}469.01 \\
(2, 1, 1, 4) & \cellcolor{lightred}OOM & \cellcolor{lightred}OOM & \cellcolor{lightred}OOM & \cellcolor{lightred}OOM & \cellcolor{lightred}OOM \\ \hline
(2, 2, 1, 1) & \cellcolor{lightgreen}396.83 & \cellcolor{lightgreen}393.96 & \cellcolor{lightgreen}391.4 & \cellcolor{lightgreen}383.28 & \cellcolor{lightgreen}382.72 \\ 
(2, 2, 1, 2) & \cellcolor{lightgreen}421.59 & \cellcolor{lightgreen}418.79 & \cellcolor{lightgreen}422.93 & \cellcolor{lightgreen}418.39 & \cellcolor{lightgreen}414.46 \\ 
(2, 2, 1, 4) & \cellcolor{lightyellow}OOM & \cellcolor{lightyellow}452.21 & \cellcolor{lightgreen}456.01 & \cellcolor{lightgreen}447.8 & \cellcolor{lightgreen}446.01 \\
(2, 2, 1, 8) & \cellcolor{lightred}OOM & \cellcolor{lightred}OOM & \cellcolor{lightred}OOM & \cellcolor{lightred}OOM & \cellcolor{lightred}OOM \\ \hline
(4, 1, 1, 1) & \cellcolor{lightgreen}408.29 & \cellcolor{lightgreen}406.13 & \cellcolor{lightgreen}402.58 & \cellcolor{lightgreen}402.04 & \cellcolor{lightgreen}397.55 \\
(4, 1, 1, 2) & \cellcolor{lightgreen}443.83 & \cellcolor{lightgreen}439.61 & \cellcolor{lightgreen}439.53 & \cellcolor{lightgreen}438.38 & \cellcolor{lightgreen}433.62 \\
(4, 1, 1, 4) & \cellcolor{lightyellow}OOM & \cellcolor{lightgreen}448.51 & \cellcolor{lightgreen}446.4 & \cellcolor{lightgreen}446.24 & \cellcolor{lightgreen}443.92 \\ \hline
(2, 1, 2, 1) & \cellcolor{lightgreen}415.39 & \cellcolor{lightgreen}413.75 & \cellcolor{lightgreen}408.00 & \cellcolor{lightgreen}410.59 & \cellcolor{lightgreen}405.03 \\
(2, 1, 2, 2) & \cellcolor{lightgreen}449.1 & \cellcolor{lightgreen}445.74 & \cellcolor{lightgreen}434.68 & \cellcolor{lightgreen}435.92 & \cellcolor{lightgreen}426.5 \\
(2, 1, 2, 4) & \cellcolor{lightred}OOM & \cellcolor{lightred}OOM & \cellcolor{lightred}OOM & \cellcolor{lightred}OOM & \cellcolor{lightred}OOM \\ \hline
(1, 2, 1, 1) & \cellcolor{lightyellow}OOM & \cellcolor{lightyellow}\textbf{495.9} & \cellcolor{lightgreen}\textbf{487.58} & \cellcolor{lightgreen}\textbf{481.69} & \cellcolor{lightgreen}\textbf{483.56} \\
(1, 2, 1, 2) & \cellcolor{lightred}OOM & \cellcolor{lightred}OOM & \cellcolor{lightred}OOM & \cellcolor{lightyellow}OOM & \cellcolor{lightyellow}OOM \\  \hline
(1, 4, 1, 1) & \cellcolor{lightyellow}\textbf{449.58} & \cellcolor{lightgreen}441.25 & \cellcolor{lightgreen}436.39 & \cellcolor{lightgreen}425.73 & \cellcolor{lightgreen}415.52 \\ 
(1, 4, 1, 2) & \cellcolor{lightyellow}OOM & \cellcolor{lightyellow}465.92 & \cellcolor{lightgreen}462.95 & \cellcolor{lightgreen}448.61 & \cellcolor{lightgreen}435.31 \\ \hline
\end{tabular}
\label{tab:h100-tflops-8192}
\end{table*}

\subsubsection{Throughput Analysis on A100(40GB)}
\label{sec:throghput-analysis-a100}

\begin{figure}[ht]
    \begin{center}
    \includegraphics[width=1.0\linewidth]{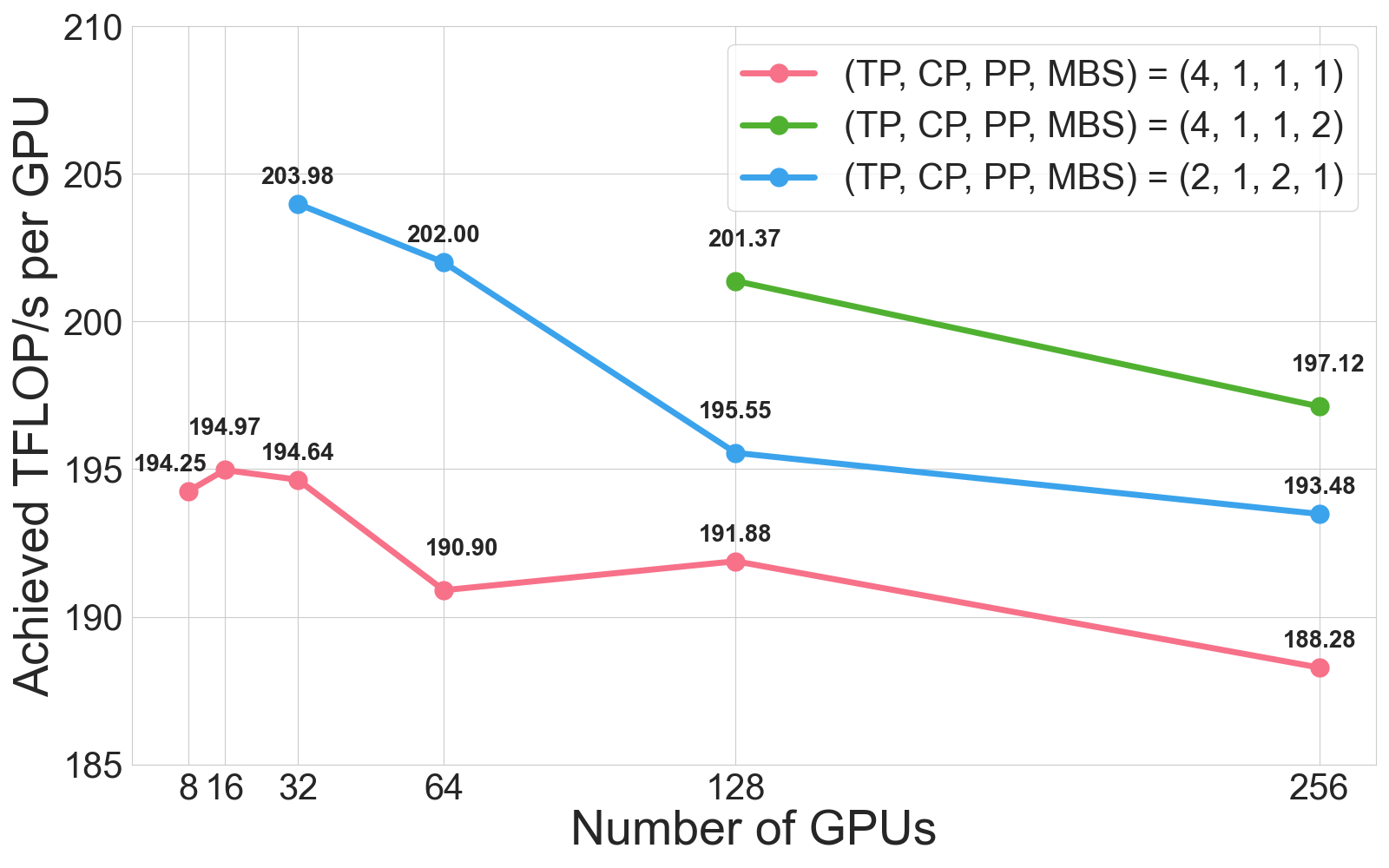}
    \end{center}
    \caption{Training TFLOP/s of Llama-3.1-8B with sequence length 8,192 on A100 (40GB). This figure illustrates how the optimal parallelism configuration changes as the number of GPUs increases.}
    \label{fig:a100-tp-pp}
\end{figure}

As shown in Table~\ref{tab:a100-tflops} and Figure~\ref{fig:a100-tp-pp}, the parallel configurations that achieve high TFLOP/s are those with the minimal combination of $TP \times CP \times PP$ that does not result in out-of-memory errors. 
This trend is also observed in Table~\ref{tab:a100-tflops-70B}, which presents the throughput for training Llama-3.1-70B in the Appendix, indicating its general applicability.

Comparing configurations using only tensor parallelism $\text{(TP, CP, PP)}=(4, 1, 1)$ with those combining tensor parallelism and pipeline parallelism $\text{(TP, CP, PP)}=(2, 1, 2)$ or tensor parallelism and context parallelism $\text{(TP, CP, PP)}=(2, 2, 1)$, we find that combining tensor parallelism with other parallelism methods achieves slightly higher TFLOP/s for the same micro batch size. 
However, since tensor parallelism can reduce both model states and activations memory, there is flexibility in choosing the micro batch size (MBS). 
In the case of $\text{(TP, CP, PP, MBS)}=(4, 1, 1, 2)$, using only tensor parallelism becomes the optimal configuration on 256GPUs.

Because context parallelism does not reduce the memory consumed by model states' parameters and gradients, it may not be possible to sufficiently distribute the optimizer states among data parallel processes and context parallel processes when using a small number of GPUs.
This can lead to out-of-memory, as observed in Table~\ref{tab:a100-tflops} for configurations with 8 to 64 GPUs.
Additionally, in the configuration $\text{(TP, CP, PP)}=(2, 1, 2)$, activation memory is not reduced in the first stage due to the nature of the 1F1B pipeline schedule~\cite{megatron-pipeline}. 
This imposes constraints on the allowable micro batch sizes to avoid out-of-memory errors.

From the above observations, we conclude that if the tensor parallel size can be kept below the number of GPUs per node, increasing the micro batch size in configurations using only tensor parallelism—up to the point where the memory consumption estimator indicates an out-of-memory condition—is a useful method for discovering sub-optimal settings.

Furthermore, in the configuration $\text{(TP, CP, PP, MBS)}=(2, 1, 2, 1)$, the number of microbatches decreases from 128 on 32 GPUs to 16 on 256 GPUs due to the increase in data parallel size.
Consequently, the pipeline bubble fraction, given by $\frac{p - 1}{m}$ where $m$ is the number of microbatches, increases by a factor of 8 (for details on the pipeline bubble fraction formula, refer to Section 2.2 of~\cite{megatron-pipeline}).
As a result, while this configuration is the fastest at 32 and 64 GPUs, its throughput decreases by approximately 6.5 TFLOP/s and 8.5 TFLOP/s on 128 and 256 GPUs, respectively, compared to the 64 GPU configuration.

These findings indicate that no specific parallelism configuration is universally optimal. However, when searching for the optimal experimental settings, it is important to avoid exploring configurations that unnecessarily increase $TP \times CP \times PP$.




As shown in Table~\ref{tab:h100-tflops-8192}, similar to the experiments on the A100 (40GB), focusing on the minimal $TP \times CP \times PP$ that does not result in out-of-memory is crucial for identifying optimal configurations.
In the case of 4 GPUs, the configuration $\text{(TP, CP, PP, MBS)}=(1, 2, 1, 1)$ results in a CUDA out-of-memory error. 
The memory consumption estimator predicts a memory usage of 89.95 GB, which is close to the 94 GB limit. This occurs because, as pointed out in Section~\ref{sec:throghput-analysis-a100}, context parallelism splits activations along the sequence dimension but does not partition the model states' parameters and gradients. 
In contrast, the configuration $\text{(TP, CP, PP, MBS)}=(2, 1, 1, 1)$ has a memory consumption of 67.52 GB on 4 GPUs according to the estimator, providing significantly more headroom.

However, while context parallelism is disadvantageous in terms of memory reduction, it has a lower communication overhead compared to tensor parallelism.
Tensor parallelism requires two all-reduce operations during forward pass per transformer layer, whereas context parallelism requires only one all-gather operation during the forward pass per transformer layer. 
Therefore, in configurations where memory is sufficient, $\text{(TP, CP, PP, MBS)}=(1, 2, 1, 1)$ is faster than $\text{(TP, CP, PP, MBS)}=(2, 1, 1, 1)$ across all comparable GPU counts from 8 to 64 GPUs.

\subsubsection{Micro Batch Size}


In the experimental results where $TP \times CP \times PP$ was not increased unnecessarily, as shown in Tables~\ref{tab:a100-tflops} and \ref{tab:h100-tflops-8192}, increasing the micro batch size consistently led to improved throughput across all configurations.
Increasing the micro batch size enhances the arithmetic intensity of executed kernels, thereby increasing GPU utilization. 
While this improves GPU utilization, it simultaneously reduces the number of microbatches. 
When using pipeline parallelism, as noted in Section~\ref{sec:throghput-analysis-a100}, this reduction can lead to increased pipeline bubbles, negatively affecting throughput.
Which effect dominates depends on factors such as the number of GPUs used and the model size, making it challenging to identify the optimal setting without experimentation. 
However, in our experiments, the positive impact of increased GPU utilization from a larger micro batch size outweighed the negative effects. 
In cases where pipeline parallelism was not used, we exclusively benefited from the increased GPU utilization, resulting in increased throughput when the micro batch size was increased.

\section{Conclusions}

In this paper, we addressed the challenge of selecting optimal parallelization configurations for large language model (LLM) training without causing GPU memory overflow.
We introduced precise formulas to estimate memory consumption when using 4D parallelism (DP, TP, PP, CP) in the Llama architecture. 
Validated through 454 experiments on NVIDIA A100 and H100 GPUs, our results confirmed that training always succeeds when estimated memory usage is below 80\% of GPU memory. 
We found that configurations minimizing $TP \times CP \times PP$ while avoiding memory overflow achieve higher throughput. Our work provides practical tools for efficiently identifying optimal configurations, enhancing resource allocation and accelerating LLM training.

\section{Reproducibility statement}

The experiments in this study were conducted using a forked code of Megatron-LM mcore v0.8.0\footnote{\url{https://github.com/NVIDIA/Megatron-LM/releases/tag/core_r0.8.0}} and TransformerEngine v1.9\footnote{\url{https://github.com/NVIDIA/TransformerEngine/releases/tag/v1.9}}.
The experimental environment included PyTorch 2.3.1+cu121, flash-attention 2.5.8, CUDA Toolkit 12.1, cuDNN 8.9.7, NCCL 2.20.5, and HPC-X 2.17.1.

\nocite{langley00}

\bibliography{example_paper}
\bibliographystyle{mlsys2024}

\clearpage
\appendix

\section{Memory Consumption Estimation and Measured TFLOP/s for Parallelism Configurations}

\subsection{A100 (40GB) Memory Consumption and TFLOP/s Measurements}

Table~\ref{tab:a100-memory-table-70b} presents the memory consumption estimates for each parallelism configuration when training Llama-3.1-70B with a sequence length of 8,192 on A100 (40GB), as predicted by the memory consumption estimator. Additionally, Table~\ref{tab:a100-tflops-70B} provides experimental results, including the observed TFLOP/s values for each configuration and whether an Out of Memory (OOM) occurred.

\begin{table}[h]
    \centering
    \caption{Estimated Total Memory (GB) per GPU when training Llama-3.1-70B on A100(40GB) with sequence length 8,192. The vertical axis (TP, CP, PP, MBS) represents Tensor Parallel size (TP), Context Parallel size (CP), Pipeline Parallel size (PP), and Micro Batch Size (MBS), respectively. Red cells indicate settings that are estimated to exceed the 40GB GPU memory limit, likely resulting in Out of Memory (OOM) errors. Yellow cells denote configurations that may approach the limit due to temporary buffers and memory fragmentation, and green cells represent estimations at or below 80\% of GPU memory (32GB or less), suggesting they are less likely to cause OOM.}
    \begin{tabular}{|c|c|c|c|}
        \hline
        (TP, CP, PP, MBS) & 64 GPUs & 128 GPUs & 256 GPUs \\
        \hline
        (8, 1, 8, 1) & \cellcolor{lightred}45.95 & \cellcolor{lightyellow}39.24 & \cellcolor{lightyellow}35.88 \\ \hline
        (8, 1, 16, 1) & - & \cellcolor{lightyellow}34.48 & \cellcolor{lightyellow}33.76 \\ \hline
        (4, 2, 8, 1) & \cellcolor{lightred}52.65 & \cellcolor{lightred}45.95 & \cellcolor{lightred}42.59 \\ \hline
        (4, 2, 16, 1) & - & \cellcolor{lightred}41.2 & \cellcolor{lightyellow}37.48 \\ \hline
        (8, 2, 8, 1) & - & \cellcolor{lightgreen}26.33 & \cellcolor{lightgreen}22.97 \\ 
        (8, 2, 8, 2) & - & \cellcolor{lightyellow}39.24 & \cellcolor{lightyellow}35.88 \\ \hline
        (8, 2, 4, 1) & \cellcolor{lightyellow}38.16 & \cellcolor{lightgreen}31.81 & \cellcolor{lightgreen}28.64 \\
        (8, 2, 4, 2) & \cellcolor{lightred}50.94 & \cellcolor{lightred}44.6 & \cellcolor{lightred}41.42 \\ \hline
        (8, 4, 4, 1) & - & \cellcolor{lightgreen}25.42 & \cellcolor{lightgreen}22.25 \\
        (8, 4, 4, 2) & - & \cellcolor{lightgreen}31.81 & \cellcolor{lightgreen}28.64 \\
        (8, 4, 4, 4) & - & \cellcolor{lightred}44.6 & \cellcolor{lightred}41.42 \\ \hline
        (8, 4, 2, 1) & \cellcolor{lightred}43.32 & \cellcolor{lightyellow}37.16 & \cellcolor{lightyellow}34.08 \\
        (8, 4, 2, 2) & \cellcolor{lightred}49.68 & \cellcolor{lightred}43.52 & \cellcolor{lightred}40.44 \\
        \hline
    \end{tabular}
    \label{tab:a100-memory-table-70b}
\end{table}

\begin{table*}[h]
    \centering
    \caption{Measured throughput(TFLOP/s) when training Llama-3.1-70B on A100 (40GB) with a sequence length of 8,192. The colors red, yellow, and green correspond to memory consumption levels predicted by the memory consumption estimator: red indicates configurations that likely exceed memory limits, yellow denotes configurations near the memory limit, and green represents configurations predicted to use 80\% or less of GPU memory (32GB or less).}
    \begin{tabular}{|c|c|c|c|}
        \hline
        (TP, CP, PP, MBS) & 64 GPUs & 128 GPUs & 256 GPUs \\
        \hline
        (8, 1, 8, 1) & \cellcolor{lightred}OOM & \cellcolor{lightyellow}OOM & \cellcolor{lightyellow}\textbf{172.31} \\ \hline
        (8, 1, 16, 1) & - & \cellcolor{lightyellow}156.74 & \cellcolor{lightyellow}149.15 \\ \hline
        (4, 2, 8, 1) & \cellcolor{lightred}OOM & \cellcolor{lightred}OOM & \cellcolor{lightred}OOM \\ \hline
        (4, 2, 16, 1) & - & \cellcolor{lightred}OOM & \cellcolor{lightyellow}157.28 \\ \hline
        (8, 2, 8, 1) & - & \cellcolor{lightgreen}156.11 & \cellcolor{lightgreen}148.02 \\
        (8, 2, 8, 2) & - & \cellcolor{lightyellow}OOM & \cellcolor{lightyellow}161.54 \\ \hline
        (8, 2, 4, 1) & \cellcolor{lightyellow}OOM & \cellcolor{lightgreen}\textbf{170.3} & \cellcolor{lightgreen}166.32 \\
        (8, 2, 4, 2) & \cellcolor{lightred}OOM & \cellcolor{lightred}OOM & \cellcolor{lightred}OOM \\ \hline
        (8, 4, 4, 1) & - & \cellcolor{lightgreen}126.6 & \cellcolor{lightgreen}123.89 \\
        (8, 4, 4, 2) & - & \cellcolor{lightgreen}157.33 & \cellcolor{lightgreen}152.53 \\
        (8, 4, 4, 4) & - & \cellcolor{lightred}OOM & \cellcolor{lightred}OOM \\ \hline
        (8, 4, 2, 1) & \cellcolor{lightred}OOM & \cellcolor{lightyellow}OOM & \cellcolor{lightyellow}132.11 \\
        (8, 4, 2, 2) & \cellcolor{lightred}OOM & \cellcolor{lightred}OOM & \cellcolor{lightred}OOM \\
        \hline
    \end{tabular}
    \label{tab:a100-tflops-70B}
\end{table*}

\subsection{H100 (94GB) Memory Consumption and TFLOP/s Measurements}

Tables~\ref{tab:h100-memory-table-8B-16384}, \ref{tab:h100-memory-table-8B-32768} show the estimated memory consumption, as predicted by the memory consumption estimator when training Llama-3.1-8B with sequence lengths of 16,386 and 32,768 on H100 (94GB). The measured TFLOP/s values for each comprehensive parallelism configuration in these settings are provided in Tables~\ref{tab:h100-tflops-8B-16384} and \ref{tab:h100-tflops-8B-32768}.

\begin{table*}[h]
    \centering
    \caption{Estimated Total Memory (GB) per GPU when training Llama-3.1-8B on H100(94GB) with sequence length 8,192. The vertical axis (TP, CP, PP, MBS) represents Tensor Parallel size (TP), Context Parallel size (CP), Pipeline Parallel size (PP), and Micro Batch Size (MBS), respectively. Red cells indicate settings that are estimated to exceed the 94GB GPU memory limit, likely resulting in Out of Memory (OOM) errors. Yellow cells denote configurations that may approach the limit due to temporary buffers and memory fragmentation, and green cells represent estimations at or below 80\% of GPU memory (75.2GB or less), suggesting they are less likely to cause OOM.}
    \begin{tabular}{|c|c|c|c|c|c|c|}
        \hline
        (TP, CP, PP, MBS) & 4 GPUs & 8 GPUs & 16 GPUs & 32 GPUs & 64 GPUs \\
        \hline
        (2, 1, 1, 1) & \cellcolor{lightgreen}67.52 & \cellcolor{lightgreen}56.30 & \cellcolor{lightgreen}50.69 & \cellcolor{lightgreen}47.89 & \cellcolor{lightgreen}46.48 \\
        (2, 1, 1, 2) & \cellcolor{lightyellow}90.16 & \cellcolor{lightyellow}78.94 & \cellcolor{lightgreen}73.34 & \cellcolor{lightgreen}70.53 & \cellcolor{lightgreen}69.13 \\
        (2, 1, 1, 4) & \cellcolor{lightred}135.45 & \cellcolor{lightred}124.23 & \cellcolor{lightred}118.62 & \cellcolor{lightred}115.82 & \cellcolor{lightred}114.42 \\ \hline
        (2, 2, 1, 1) & \cellcolor{lightgreen}56.20 & \cellcolor{lightgreen}44.98 & \cellcolor{lightgreen}39.37 & \cellcolor{lightgreen}36.56 & \cellcolor{lightgreen}35.16 \\
        (2, 2, 1, 2) & \cellcolor{lightgreen}67.52 & \cellcolor{lightgreen}56.30 & \cellcolor{lightgreen}50.69 & \cellcolor{lightgreen}47.89 & \cellcolor{lightgreen}46.48 \\
        (2, 2, 1, 4) & \cellcolor{lightyellow}90.16 & \cellcolor{lightyellow}78.94 & \cellcolor{lightgreen}73.33 & \cellcolor{lightgreen}70.53 & \cellcolor{lightgreen}69.13 \\
        (2, 2, 1, 8) & \cellcolor{lightred}135.45 & \cellcolor{lightred}124.23 & \cellcolor{lightred}118.62 & \cellcolor{lightred}115.82 & \cellcolor{lightred}114.42 \\ \hline
        (4, 1, 1, 1) & \cellcolor{lightgreen}44.98 & \cellcolor{lightgreen}33.76 & \cellcolor{lightgreen}28.15 & \cellcolor{lightgreen}25.35 & \cellcolor{lightgreen}23.94 \\
        (4, 1, 1, 2) & \cellcolor{lightgreen}56.30 & \cellcolor{lightgreen}45.08 & \cellcolor{lightgreen}39.47 & \cellcolor{lightgreen}36.67 & \cellcolor{lightgreen}35.27 \\
        (4, 1, 1, 4) & \cellcolor{lightyellow}78.95 & \cellcolor{lightgreen}67.73 & \cellcolor{lightgreen}62.12 & \cellcolor{lightgreen}59.31 & \cellcolor{lightgreen}57.91 \\ \hline
        (2, 1, 2, 1) & \cellcolor{lightgreen}54.41 & \cellcolor{lightgreen}43.19 & \cellcolor{lightgreen}37.58 & \cellcolor{lightgreen}34.77 & \cellcolor{lightgreen}33.37 \\
        (2, 1, 2, 2) & \cellcolor{lightgreen}75.16 & \cellcolor{lightgreen}63.94 & \cellcolor{lightgreen}58.33 & \cellcolor{lightgreen}55.52 & \cellcolor{lightgreen}54.12 \\
        (2, 1, 2, 4) & \cellcolor{lightred}116.66 & \cellcolor{lightred}105.43 & \cellcolor{lightred}99.83 & \cellcolor{lightred}97.02 & \cellcolor{lightred}95.62 \\ \hline
        (1, 2, 1, 1) & \cellcolor{lightyellow}89.95 & \cellcolor{lightyellow}78.74 & \cellcolor{lightgreen}70.32 & \cellcolor{lightgreen}68.92 & \cellcolor{lightgreen}68.22 \\
        (1, 2, 1, 2) & \cellcolor{lightred}112.60 & \cellcolor{lightred}101.38 & \cellcolor{lightred}95.77 & \cellcolor{lightyellow}92.97 & \cellcolor{lightyellow}91.56 \\ \hline
        (1, 4, 1, 1) & \cellcolor{lightyellow}78.63 & \cellcolor{lightgreen}67.41 & \cellcolor{lightgreen}61.80 & \cellcolor{lightgreen}59.00 & \cellcolor{lightgreen}57.60 \\
        (1, 4, 1, 2) & \cellcolor{lightyellow}89.95 & \cellcolor{lightyellow}78.74 & \cellcolor{lightgreen}73.13 & \cellcolor{lightgreen}70.32 & \cellcolor{lightgreen}68.92 \\
        \hline
    \end{tabular}
    \label{tab:h100-memory-table-8B-8192}
\end{table*}

\begin{table*}[h]
    \centering
    \caption{Estimated Total Memory (GB) per GPU when training Llama-3.1-8B on H100(94GB) with sequence length 16,384. The vertical axis (TP, CP, PP, MBS) represents Tensor Parallel size (TP), Context Parallel size (CP), Pipeline Parallel size (PP), and Micro Batch Size (MBS), respectively. Red cells indicate settings that are estimated to exceed the 94GB GPU memory limit, likely resulting in Out of Memory (OOM) errors. Yellow cells denote configurations that may approach the limit due to temporary buffers and memory fragmentation, and green cells represent estimations at or below 80\% of GPU memory (75.2GB or less), suggesting they are less likely to cause OOM.}
    \begin{tabular}{|c|c|c|c|c|c|}
        \hline
        (TP, CP, PP, MBS) & 4 GPUs & 8 GPUs & 16 GPUs & 32 GPUs & 64 GPUs \\
        \hline
        (2, 1, 1, 1) & \cellcolor{lightyellow}90.16 & \cellcolor{lightyellow}78.94 & \cellcolor{lightgreen}73.34 & \cellcolor{lightgreen}70.53 & \cellcolor{lightgreen}69.13 \\
        (2, 1, 1, 2) & \cellcolor{lightred}135.45 & \cellcolor{lightred}124.23 & \cellcolor{lightred}118.62 & \cellcolor{lightred}115.82 & \cellcolor{lightred}114.42 \\
        (2, 1, 1, 4) & \cellcolor{lightred}226.03 & \cellcolor{lightred}214.81 & \cellcolor{lightred}209.20 & \cellcolor{lightred}206.40 & \cellcolor{lightred}205.00 \\ \hline
        (2, 2, 1, 1) & \cellcolor{lightgreen}67.52 & \cellcolor{lightgreen}56.30 & \cellcolor{lightgreen}50.69 & \cellcolor{lightgreen}47.89 & \cellcolor{lightgreen}46.48 \\
        (2, 2, 1, 2) & \cellcolor{lightyellow}90.16 & \cellcolor{lightyellow}78.94 & \cellcolor{lightgreen}73.34 & \cellcolor{lightgreen}70.53 & \cellcolor{lightgreen}69.13 \\
        (2, 2, 1, 4) & \cellcolor{lightred}135.45 & \cellcolor{lightred}124.23 & \cellcolor{lightred}118.62 & \cellcolor{lightred}115.82 & \cellcolor{lightred}114.42 \\
        (2, 2, 1, 8) & \cellcolor{lightred}226.03 & \cellcolor{lightred}214.81 & \cellcolor{lightred}209.20 & \cellcolor{lightred}206.4 & \cellcolor{lightred}205.00 \\ \hline
        (4, 1, 1, 1) & \cellcolor{lightgreen}56.30 & \cellcolor{lightgreen}45.08 & \cellcolor{lightgreen}39.47 & \cellcolor{lightgreen}36.67 & \cellcolor{lightgreen}35.27 \\
        (4, 1, 1, 2) & \cellcolor{lightyellow}78.95 & \cellcolor{lightgreen}67.73 & \cellcolor{lightgreen}62.12 & \cellcolor{lightgreen}59.31 & \cellcolor{lightgreen}57.91 \\
        (4, 1, 1, 4) & \cellcolor{lightred}124.24 & \cellcolor{lightred}113.02 & \cellcolor{lightred}107.41 & \cellcolor{lightred}104.6 & \cellcolor{lightred}103.2 \\ \hline
        (2, 1, 2, 1) & \cellcolor{lightgreen}75.16 & \cellcolor{lightgreen}63.94 & \cellcolor{lightgreen}58.33 & \cellcolor{lightgreen}55.52 & \cellcolor{lightgreen}54.12 \\
        (2, 1, 2, 2) & \cellcolor{lightred}116.66 & \cellcolor{lightred}105.44 & \cellcolor{lightred}99.83 & \cellcolor{lightred}97.02 & \cellcolor{lightred}95.62 \\
        (2, 1, 2, 4) & \cellcolor{lightred}199.66 & \cellcolor{lightred}188.44 & \cellcolor{lightred}182.83 & \cellcolor{lightred}180.02 & \cellcolor{lightred}178.62 \\ \hline
        (1, 2, 1, 1) & \cellcolor{lightred}112.6 & \cellcolor{lightred}101.38 & \cellcolor{lightred}95.77 & \cellcolor{lightyellow}92.97 & \cellcolor{lightyellow}91.56 \\
        (1, 2, 1, 2) & \cellcolor{lightred}157.89 & \cellcolor{lightred}146.67 & \cellcolor{lightred}141.06 & \cellcolor{lightred}138.26 & \cellcolor{lightred}136.85 \\ \hline
        (1, 4, 1, 1) & \cellcolor{lightyellow}89.95 & \cellcolor{lightyellow}78.74 & \cellcolor{lightgreen}73.13 & \cellcolor{lightgreen}70.32 & \cellcolor{lightgreen}68.92 \\
        (1, 4, 1, 2) & \cellcolor{lightred}112.60 & \cellcolor{lightred}101.38 & \cellcolor{lightred}95.77 & \cellcolor{lightyellow}92.97 & \cellcolor{lightyellow}91.56 \\
        (1, 4, 1, 4) & \cellcolor{lightred}157.89 & \cellcolor{lightred}146.67 & \cellcolor{lightred}141.06 & \cellcolor{lightred}138.25 & \cellcolor{lightred}136.85 \\ \hline
        (2, 4, 1, 1) & - & \cellcolor{lightgreen}44.98 & \cellcolor{lightgreen}39.37 & \cellcolor{lightgreen}36.56 & \cellcolor{lightgreen}35.16 \\
        (2, 4, 1, 2) & - & \cellcolor{lightgreen}56.30 & \cellcolor{lightgreen}50.69 & \cellcolor{lightgreen}47.89 & \cellcolor{lightgreen}46.48 \\ \hline
        (4, 2, 1, 1) & - & \cellcolor{lightgreen}33.76 & \cellcolor{lightgreen}28.15 & \cellcolor{lightgreen}25.35 & \cellcolor{lightgreen}23.94 \\
        (4, 2, 1, 2) & - & \cellcolor{lightgreen}45.08 & \cellcolor{lightgreen}39.47 & \cellcolor{lightgreen}36.67 & \cellcolor{lightgreen}35.27 \\
        \hline
    \end{tabular}
    \label{tab:h100-memory-table-8B-16384}
\end{table*}

\begin{table*}[h]
    \centering
    \caption{Estimated Total Memory (GB) per GPU when training Llama-3.1-8B on H100(94GB) with sequence length 32,768. The vertical axis (TP, CP, PP, MBS) represents Tensor Parallel size (TP), Context Parallel size (CP), Pipeline Parallel size (PP), and Micro Batch Size (MBS), respectively. Red cells indicate settings that are estimated to exceed the 94GB GPU memory limit, likely resulting in Out of Memory (OOM) errors. Yellow cells denote configurations that may approach the limit due to temporary buffers and memory fragmentation, and green cells represent estimations at or below 80\% of GPU memory (75.2GB or less), suggesting they are less likely to cause OOM.}
    \begin{tabular}{|c|c|c|c|c|c|}
        \hline
        (TP, CP, PP, MBS) & 4 GPUs & 8 GPUs & 16 GPUs & 32 GPUs & 64 GPUs \\
        \hline
        (2, 1, 1, 1) & \cellcolor{lightred}135.45 & \cellcolor{lightred}124.23 & \cellcolor{lightred}118.62 & \cellcolor{lightred}115.82 & \cellcolor{lightred}114.42 \\
        (2, 1, 1, 2) & \cellcolor{lightred}226.03 & \cellcolor{lightred}214.81 & \cellcolor{lightred}209.20 & \cellcolor{lightred}206.40 & \cellcolor{lightred}205.00 \\
        (2, 1, 1, 4) & \cellcolor{lightred}407.19 & \cellcolor{lightred}396.97 & \cellcolor{lightred}390.36 & \cellcolor{lightred}387.55 & \cellcolor{lightred}386.15 \\ \hline
        (2, 2, 1, 1) & \cellcolor{lightyellow}90.16 & \cellcolor{lightyellow}78.94 & \cellcolor{lightgreen}73.34 & \cellcolor{lightgreen}70.53 & \cellcolor{lightgreen}69.13 \\
        (2, 2, 1, 2) & \cellcolor{lightred}135.45 & \cellcolor{lightred}124.23 & \cellcolor{lightred}118.62 & \cellcolor{lightred}115.82 & \cellcolor{lightred}114.42 \\
        (2, 2, 1, 4) & \cellcolor{lightred}226.03 & \cellcolor{lightred}214.81 & \cellcolor{lightred}209.2 & \cellcolor{lightred}206.40 & \cellcolor{lightred}205.00 \\
        (2, 2, 1, 8) & \cellcolor{lightred}407.19 & \cellcolor{lightred}395.97 & \cellcolor{lightred}390.36 & \cellcolor{lightred}387.55 & \cellcolor{lightred}386.15 \\ \hline
        (4, 1, 1, 1) & \cellcolor{lightyellow}78.95 & \cellcolor{lightgreen}67.73 & \cellcolor{lightgreen}62.12 & \cellcolor{lightgreen}59.31 & \cellcolor{lightgreen}57.91 \\
        (4, 1, 1, 2) & \cellcolor{lightred}124.24 & \cellcolor{lightred}113.02 & \cellcolor{lightred}107.41 & \cellcolor{lightred}104.60 & \cellcolor{lightred}103.20 \\
        (4, 1, 1, 4) & \cellcolor{lightred}214.81 & \cellcolor{lightred}203.59 & \cellcolor{lightred}197.99 & \cellcolor{lightred}195.18 & \cellcolor{lightred}193.78 \\ \hline
        (2, 1, 2, 1) & \cellcolor{lightred}116.66 & \cellcolor{lightred}105.44 & \cellcolor{lightred}99.83 & \cellcolor{lightred}97.02 & \cellcolor{lightred}95.62 \\ \hline
        (1, 4, 1, 1) & \cellcolor{lightred}112.6 & \cellcolor{lightred}101.38 & \cellcolor{lightred}95.77 & \cellcolor{lightyellow}92.97 & \cellcolor{lightyellow}91.56 \\ \hline
        (2, 4, 1, 1) & - & \cellcolor{lightgreen}56.30 & \cellcolor{lightgreen}50.69 & \cellcolor{lightgreen}47.89 & \cellcolor{lightgreen}46.48 \\
        (2, 4, 1, 2) & - & \cellcolor{lightyellow}78.94 & \cellcolor{lightgreen}73.34 & \cellcolor{lightgreen}70.53 & \cellcolor{lightgreen}69.13 \\ \hline
        (4, 2, 1, 1) & - & \cellcolor{lightgreen}45.08 & \cellcolor{lightgreen}39.47 & \cellcolor{lightgreen}36.67 & \cellcolor{lightgreen}35.27 \\
        (4, 2, 1, 2) & - & \cellcolor{lightgreen}67.73 & \cellcolor{lightgreen}62.12 & \cellcolor{lightgreen}59.31 & \cellcolor{lightgreen}57.91 \\ \hline
        (2, 2, 2, 1) & - & \cellcolor{lightgreen}63.94 & \cellcolor{lightgreen}58.33 & \cellcolor{lightgreen}55.52 & \cellcolor{lightgreen}54.12 \\ \hline
        (1, 4, 2, 1) & - & \cellcolor{lightgreen}75.15 & \cellcolor{lightgreen}69.55 & \cellcolor{lightgreen}66.74 & \cellcolor{lightgreen}65.34 \\
        \hline
    \end{tabular}
    \label{tab:h100-memory-table-8B-32768}
\end{table*}

\begin{table*}[h]
    \centering
    \caption{Measured throughput(TFLOP/s) when training Llama-3.1-8B on H100 (94GB) with a sequence length of 16,384. The colors red, yellow, and green correspond to memory consumption levels predicted by the memory consumption estimator: red indicates configurations that likely exceed memory limits, yellow denotes configurations near the memory limit, and green represents configurations predicted to use 80\% or less of GPU memory (75.2GB or less).}
    \begin{tabular}{|c|c|c|c|c|c|}
        \hline
        (TP, CP, PP, MBS) & 4 GPUs & 8 GPUs & 16 GPUs & 32 GPUs & 64 GPUs \\
        \hline
        (2, 1, 1, 1) & \cellcolor{lightyellow}OOM & \cellcolor{lightyellow}497.56 & \cellcolor{lightgreen}494.06 & \cellcolor{lightgreen}493.34 & \cellcolor{lightgreen}492.28 \\
        (2, 1, 1, 2) & \cellcolor{lightred}OOM & \cellcolor{lightred}OOM & \cellcolor{lightred}OOM & \cellcolor{lightred}OOM & \cellcolor{lightred}OOM \\
        (2, 1, 1, 4) & \cellcolor{lightred}OOM & \cellcolor{lightred}OOM & \cellcolor{lightred}OOM & \cellcolor{lightred}OOM & \cellcolor{lightred}OOM \\ \hline
        (2, 2, 1, 1) & \cellcolor{lightgreen}425.29 & \cellcolor{lightgreen}432.15 & \cellcolor{lightgreen}420.1 & \cellcolor{lightgreen}388.7 & \cellcolor{lightgreen}365.75 \\
        (2, 2, 1, 2) & \cellcolor{lightyellow}453.76 & \cellcolor{lightyellow}463.88 & \cellcolor{lightgreen}459.94 & \cellcolor{lightgreen}452.71 & \cellcolor{lightgreen}436.77 \\
        (2, 2, 1, 4) & \cellcolor{lightred}OOM & \cellcolor{lightred}OOM & \cellcolor{lightred}OOM & \cellcolor{lightred}OOM & \cellcolor{lightred}OOM \\
        (2, 2, 1, 8) & \cellcolor{lightred}OOM & \cellcolor{lightred}OOM & \cellcolor{lightred}OOM & \cellcolor{lightred}OOM & \cellcolor{lightred}OOM \\ \hline
        (4, 1, 1, 1) & \cellcolor{lightgreen}442.54 & \cellcolor{lightgreen}448.18 & \cellcolor{lightgreen}438.37 & \cellcolor{lightgreen}406.59 & \cellcolor{lightgreen}384.14 \\
        (4, 1, 1, 2) & \cellcolor{lightyellow}451.10 & \cellcolor{lightgreen}468.49 & \cellcolor{lightgreen}462.28 & \cellcolor{lightgreen}453.71 & \cellcolor{lightgreen}457.36 \\
        (4, 1, 1, 4) & \cellcolor{lightred}OOM & \cellcolor{lightred}OOM & \cellcolor{lightred}OOM & \cellcolor{lightred}OOM & \cellcolor{lightred}OOM \\ \hline
        (2, 1, 2, 1) & \cellcolor{lightgreen}448.13 & \cellcolor{lightgreen}450.76 & \cellcolor{lightgreen}433.43 & \cellcolor{lightgreen}399.61 & \cellcolor{lightgreen}365.75 \\
        (2, 1, 2, 2) & \cellcolor{lightred}OOM & \cellcolor{lightred}OOM & \cellcolor{lightred}OOM & \cellcolor{lightred}OOM & \cellcolor{lightred}OOM \\
        (2, 1, 2, 4) & \cellcolor{lightred}OOM & \cellcolor{lightred}OOM & \cellcolor{lightred}OOM & \cellcolor{lightred}OOM & \cellcolor{lightred}OOM \\ \hline
        (1, 2, 1, 1) & \cellcolor{lightred}OOM & \cellcolor{lightred}OOM & \cellcolor{lightred}OOM & \cellcolor{lightyellow}OOM & \cellcolor{lightyellow}OOM \\
        (1, 2, 1, 2) & \cellcolor{lightred}OOM & \cellcolor{lightred}OOM & \cellcolor{lightred}OOM & \cellcolor{lightred}OOM & \cellcolor{lightred}OOM \\ \hline
        (1, 4, 1, 1) & \cellcolor{lightyellow}OOM & \cellcolor{lightyellow}461.64 & \cellcolor{lightgreen}439.89 & \cellcolor{lightgreen}409.41 & \cellcolor{lightgreen}363.59 \\
        (1, 4, 1, 2) & \cellcolor{lightred}OOM & \cellcolor{lightred}OOM & \cellcolor{lightred}OOM & \cellcolor{lightyellow}OOM & \cellcolor{lightyellow}OOM \\
        (1, 4, 1, 4) & \cellcolor{lightred}OOM & \cellcolor{lightred}OOM & \cellcolor{lightred}OOM & \cellcolor{lightred}OOM & \cellcolor{lightred}OOM \\ \hline
        (2, 4, 1, 1) & - & \cellcolor{lightgreen}288.42 & \cellcolor{lightgreen}290.7 & \cellcolor{lightgreen}267.03 & \cellcolor{lightgreen}237.84 \\
        (2, 4, 1, 2) & - & \cellcolor{lightgreen}272.96 & \cellcolor{lightgreen}293.26 & \cellcolor{lightgreen}279.53 & \cellcolor{lightgreen}273.22 \\ \hline
        (4, 2, 1, 1) & - & \cellcolor{lightgreen}303.46 & \cellcolor{lightgreen}295.71 & \cellcolor{lightgreen}295.71 & \cellcolor{lightgreen}255.45 \\ 
        (4, 2, 1, 2) & - & \cellcolor{lightgreen}297.73 & \cellcolor{lightgreen}299.95 & \cellcolor{lightgreen}299.95 & \cellcolor{lightgreen}282.24 \\
        \hline
    \end{tabular}
    \label{tab:h100-tflops-8B-16384}
\end{table*}

\begin{table*}[h]
    \centering
    \caption{Measured throughput(TFLOP/s) when training Llama-3.1-8B on H100 (94GB) with a sequence length of 32,768. The colors red, yellow, and green correspond to memory consumption levels predicted by the memory consumption estimator: red indicates configurations that likely exceed memory limits, yellow denotes configurations near the memory limit, and green represents configurations predicted to use 80\% or less of GPU memory (75.2GB or less).}
    \begin{tabular}{|c|c|c|c|c|c|}
        \hline
        (TP, CP, PP, MBS) & 4 GPUs & 8 GPUs & 16 GPUs & 32 GPUs & 64 GPUs \\
        \hline
        (2, 1, 1, 1) & \cellcolor{lightred}OOM & \cellcolor{lightred}OOM & \cellcolor{lightred}OOM & \cellcolor{lightred}OOM & \cellcolor{lightred}OOM \\
        (2, 1, 1, 2) & \cellcolor{lightred}OOM & \cellcolor{lightred}OOM & \cellcolor{lightred}OOM & \cellcolor{lightred}OOM & \cellcolor{lightred}OOM \\
        (2, 1, 1, 4) & \cellcolor{lightred}OOM & \cellcolor{lightred}OOM & \cellcolor{lightred}OOM & \cellcolor{lightred}OOM & \cellcolor{lightred}OOM \\ \hline
        (2, 2, 1, 1) & \cellcolor{lightyellow}OOM & \cellcolor{lightyellow}336.94 & \cellcolor{lightgreen}308.35 & \cellcolor{lightgreen}288.00 & \cellcolor{lightgreen}229.30 \\
        (2, 2, 1, 2) & \cellcolor{lightred}OOM & \cellcolor{lightred}OOM & \cellcolor{lightred}OOM & \cellcolor{lightred}OOM & \cellcolor{lightred}OOM \\
        (2, 2, 1, 4) & \cellcolor{lightred}OOM & \cellcolor{lightred}OOM & \cellcolor{lightred}OOM & \cellcolor{lightred}OOM & \cellcolor{lightred}OOM \\
        (2, 2, 1, 8) & \cellcolor{lightred}OOM & \cellcolor{lightred}OOM & \cellcolor{lightred}OOM & \cellcolor{lightred}OOM & \cellcolor{lightred}OOM \\ \hline
        (4, 1, 1, 1) & \cellcolor{lightyellow}363.68 & \cellcolor{lightgreen}340.51 & \cellcolor{lightgreen}318.83 & \cellcolor{lightgreen}290.18 & \cellcolor{lightgreen}234.85 \\
        (4, 1, 1, 2) & \cellcolor{lightred}OOM & \cellcolor{lightred}OOM & \cellcolor{lightred}OOM & \cellcolor{lightred}OOM & \cellcolor{lightred}OOM \\
        (4, 1, 1, 4) & \cellcolor{lightred}OOM & \cellcolor{lightred}OOM & \cellcolor{lightred}OOM & \cellcolor{lightred}OOM & \cellcolor{lightred}OOM \\ \hline
        (2, 1, 2, 1) & \cellcolor{lightred}OOM & \cellcolor{lightred}OOM & \cellcolor{lightred}OOM & \cellcolor{lightred}OOM & \cellcolor{lightred}OOM \\ \hline
        (1, 4, 1, 1) & \cellcolor{lightred}OOM & \cellcolor{lightred}OOM & \cellcolor{lightred}OOM & \cellcolor{lightyellow}OOM & \cellcolor{lightyellow}OOM \\ \hline
        (2, 4, 1, 1) & - & \cellcolor{lightgreen}179.97 & \cellcolor{lightgreen}170.11 & \cellcolor{lightgreen}152.09 & \cellcolor{lightgreen}135.51 \\
        (2, 4, 1, 2) & - & \cellcolor{lightyellow}178.79 & \cellcolor{lightgreen}173.15 & \cellcolor{lightgreen}162.30 & \cellcolor{lightgreen}151.39 \\ \hline
        (4, 2, 1, 1) & - & \cellcolor{lightgreen}181.36 & \cellcolor{lightgreen}174.21 & \cellcolor{lightgreen}156.90 & \cellcolor{lightgreen}146.43 \\
        (4, 2, 1, 2) & - & \cellcolor{lightgreen}181.56 & \cellcolor{lightgreen}176.28 & \cellcolor{lightgreen}162.30 & \cellcolor{lightgreen}145.65 \\ \hline
        (2, 2, 2, 1) & - & \cellcolor{lightgreen}178.22 & \cellcolor{lightgreen}170.10 & \cellcolor{lightgreen}154.70 & \cellcolor{lightgreen}142.97 \\ \hline
        (1, 4, 2, 1) & - & \cellcolor{lightgreen}169.49 & \cellcolor{lightgreen}170.70 & \cellcolor{lightgreen}151.90 & \cellcolor{lightgreen}131.50 \\
        \hline
    \end{tabular}
    \label{tab:h100-tflops-8B-32768}
\end{table*}

%


\end{document}